\documentclass[nohyperref]{article}

\usepackage{microtype}
\usepackage{graphicx}
\usepackage{booktabs} 

\usepackage{hyperref}

\usepackage[accepted]{icml2022_arxiv}

\usepackage{amsmath}
\usepackage{amssymb}
\usepackage{mathtools}
\usepackage{amsthm}

\usepackage{pifont}
\usepackage{commath}
\usepackage{mathtools}
\usepackage{subcaption}
\usepackage{algorithm}
\usepackage{algorithmic}
\usepackage{subcaption}
\usepackage{multirow}
\usepackage[flushleft]{threeparttable}
\usepackage{xcolor}
\let\norm\undefined 
\DeclarePairedDelimiter\norm{\lVert}{\rVert}

\definecolor{darkergreen}{RGB}{21, 152, 56}
\definecolor{red2}{RGB}{252, 54, 65}

\newcommand{\cmark}{\textcolor{darkergreen}{\ding{52}}}%
\newcommand{\xmark}{\textcolor{red2}{\ding{56}}}%
\usepackage[capitalize,noabbrev]{cleveref}

\theoremstyle{plain}

\theoremstyle{definition}

\theoremstyle{remark}

\usepackage[textsize=tiny]{todonotes}

\icmltitlerunning{Task-Adaptive Feature Transformer with Semantic Enrichment for Few-Shot Segmentation}

\begin{document}
	
	\twocolumn[
	\icmltitle{Task-Adaptive Feature Transformer with Semantic Enrichment \\for Few-Shot Segmentation}
	
	
	
	\icmlsetsymbol{equal}{*}
	
	\begin{icmlauthorlist}
		\icmlauthor{Jun Seo}{kaist}
		\icmlauthor{Young-Hyun Park}{kaist}
		\icmlauthor{Sung Whan Yoon}{unist}
		\icmlauthor{Jaekyun Moon}{kaist}
	\end{icmlauthorlist}
	
	\icmlaffiliation{kaist}{School of Electrical Engineering, Korea Advanced Institure of Science and Technology (KAIST), Daejeon, Korea}
	\icmlaffiliation{unist}{School of Electrical and Computer Engineering, Ulsan National Institute of Science and Technology (UNIST), Ulsan, Korea}
	
	\icmlcorrespondingauthor{Jun Seo}{tjwns0630@kaist.ac.kr}

	\icmlkeywords{Machine Learning, ICML}
	
	\vskip 0.3in
	]
	
	
	
	\printAffiliationsAndNotice{\icmlEqualContribution} 
	
	\begin{abstract}
		Few-shot learning allows machines to classify novel classes using only a few labeled samples. Recently, few-shot segmentation aiming at semantic segmentation on low sample data has also seen great interest. In this paper, we propose a learnable module that can be placed on top of existing segmentation networks for performing few-shot segmentation. This module, called the task-adaptive feature transformer (TAFT), linearly transforms task-specific high-level features to a set of task agnostic features well-suited to conducting few-shot segmentation. The task-conditioned feature transformation allows an effective utilization of the semantic information in novel classes to generate tight segmentation masks. We also propose a semantic enrichment (SE) module that utilizes a pixel-wise attention module for high-level feature and an auxiliary loss from an auxiliary segmentation network conducting the semantic segmentation for all training classes.
		Experiments on PASCAL-$5^i$ and COCO-$20^i$ datasets confirm that the added modules successfully extend the capability of existing segmentators to yield highly competitive few-shot segmentation performances.
		
	\end{abstract}
	\section{Introduction}
	\vspace{-2mm}
	Deep neural networks have made significant advances in computer vision tasks such as image classification \cite{ Alexnet, googlenet, inception, Resnet}, object detection \cite{FasterRCNN, YOLO}, and semantic segmentation \cite{FCN, UNet,PSPNet, DeeplabV1, DeeplabV3, DeeplabV3+}. However, training a deep neural network requires a large amount of labeled data, which are scarce or expensive in many cases. Few-shot learning algorithms aim to tackle this problem. Advances in few-shot learning allows machines to handle previously unseen classification tasks with only a few labeled samples in some cases \cite{MN,MAML,PN,TapNet}.
	
	Recently, more complicated few-shot learning problems such as few-shot object detection \cite{FSD1, FSD2, FSD3} and few-shot semantic segmentation \cite{OSLSM, coFCN, CANet} have seen much interest. Labels for object detection or semantic segmentation are even harder to obtain, naturally occasioning the formulation of few-shot detection and few-shot segmentation problems. 
	In this paper, we tackle the few-shot semantic segmentation problem. The goal of few-shot segmentation is to conduct semantic segmentation on previously unseen classes with only a few examples. Few-shot segmentation tends to be even more challenging than few-shot classification, since segmentation labels contain more information than classification labels.
	
	Specifically, we propose a learnable module, a task-adaptive feature transformer (TAFT), that can extend 
	existing segmentation networks to acquire few-shot segmentation capability. TAFT can be plugged into any existing semantic segmentation algorithm employing an encoder-decoder structure.
	In encoder-decoder-based segmentators, the encoder extracts features from input images, and the decoder generates segmentation masks using the features. Although multi-scale features can be utilized, it is the high level features that contain the most semantic information. The proposed TAFT method is able to convert the semantic information from novel classes into a form that is more comprehensible to the decoder, by transforming the high-level features to task-agnostic features that contain the information sufficient for segmentation regardless of the given task.	
	Few-shot segmentation can be done effectively by the decoder generating the segmentation mask based on these task-agnostic features.
	
	The transformation matrix in TAFT, which is the only moving part that changes according to the given task, realizes the linear transformation that brings all the class prototypes in the embedded feature space close to a set of task-independent class references in the task-agnostic feature space.
	These task-independent reference vectors are meta-learned with the update taking place at the end of every episode processing stage, but are completely decoupled from the current task. The references are shown to provide a highly stable target point for the linear transformation, and the decoder more easily generates the segmentation mask based on the transformed task-agnostic features.  
	
	Since the feature transformation of TAFT relies on the high-level features extracted from the encoder, it is essential to obtain the informative high-level features in the first place. To improve the quality of the high-level features, we propose a semantic enrichment (SE) method using following techniques. First, we adopt a pixel-wise self-attention module to enhance the quality of the semantic information in the high-level features by reflecting the context among feature pixels. Second, we use an auxiliary segmentation loss from an auxiliary decoder that conducts the multi-class semantic segmentation using the encoder features. By doing so, the encoder can learn to extract the more general high-level features which are advantageous for the few-shot adaptation to the novel classes.
	
	Because our TAFT and SE are simple add-on modules, we do not expect the extended segmentators to perform as well as the tailored state-of-the-art (SOTA) few-shot segmentators designed from the scratch, on extremely low-shot cases like one-shot. Our goal, however, is to see that as the number of shots increases to a few shots, e.g., 5 shots, the TAFT-SE add-on modules would learn quickly and allow the existing segmentator to compete strongly with fully-optimized SOTA few-shot segmentation models.   
	
	We evaluate the few-shot learning capability of the proposed modules by combining them with the well-known segmentation algorithm, Deeplab V3+. We indeed observe that while the resulting one-shot performance is not SOTA, for 5-shot testing on both PASCAL-$5^i$ and COCO-$20^i$ datasets, TAFT-SE plugged into Deeplab V3+ beats the SOTA few-shot segmentation algorithms by substantial margins. 
	\vspace{-3mm}
	\section{Proposed Method}
	\vspace{-1mm}
	\subsection{Problem Definition}
	\vspace{-2mm}
	The goal of few-shot segmentation is to meta-train the model so that it can adapt to unseen classes and perform segmentation with only a few labeled samples from new classes. For meta-training and evaluation, training set $D_{train}$ and test set $D_{test}$ with no-overlapping categories are used. In both meta-training and evaluation phases, episodes are composed with classes randomly selected from $D_{train}$ and $D_{test}$, respectively. For 1-way $N$-shot segmentation, each episode consists of a support set $S=\{(\mathbf{x}_{n}, \mathbf{y}_{n})\}_{n=1}^{N}$ with $N$ image-label pairs and a query set $Q=\{(\bar{\mathbf{x}}_{m}, \bar{\mathbf{y}}_{m})\}_{m=1}^{M}$ with $M$ image-label pairs, where $\mathbf{x}_{n}$ and $\bar{\mathbf{x}}_{m}$ represent images and $\mathbf{y}_{n}$ and $\bar{\mathbf{y}}_{m}$ indicate the corresponding binary masks for the given class. In each episode, the model adapts to the given class using $S$, and then predicts the segmentation masks from the query images $\{\bar{\mathbf{x}}_{m}\}_{m=1}^{M}$. At the end of each episode, the predicted segmentation masks are compared with the query masks $\{\bar{\mathbf{y}}_{m}\}_{m=1}^{M}$. In the meta-training phase, a loss is computed based on the difference between the predicted masks and the query masks and used to update the model parameters. In the evaluation phase, the model performance is evaluated from the comparison between the predicted masks and the query masks.
	\vspace{-3mm}
	\subsection{Task-Adaptive Feature Transformer}
	\label{section2_1}
\begin{figure*}[h]
		\centering
		\includegraphics[width=0.8\textwidth]{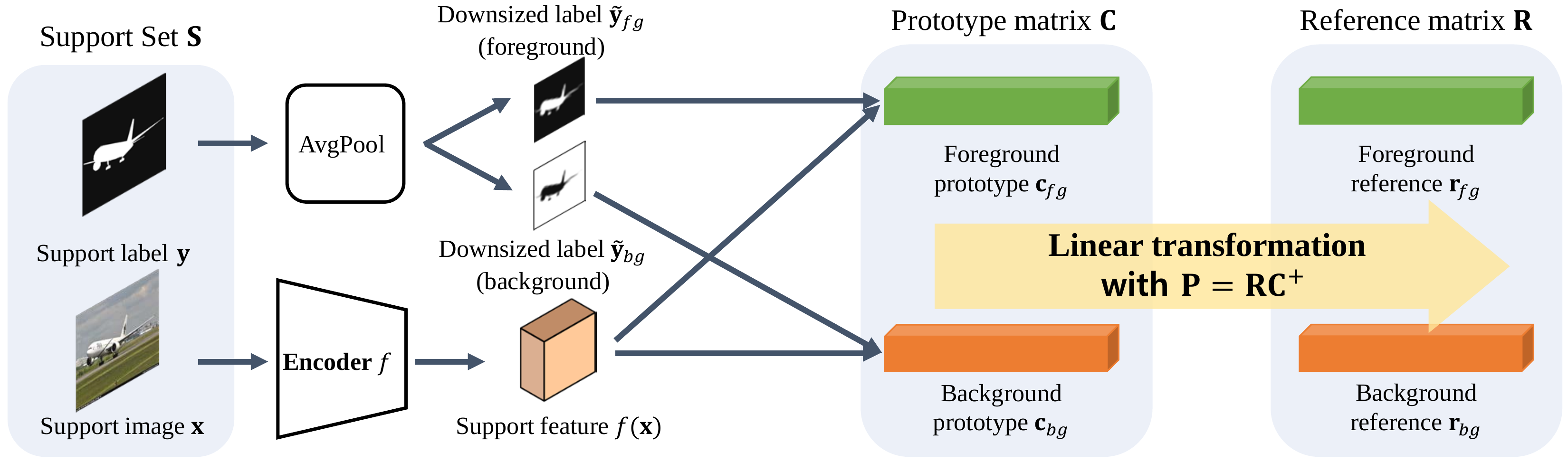}
		\vspace{-2mm}
		\caption{Process of constructing transformation matrix in TAFT}
		\label{figure1}
	\end{figure*}

	\vspace{-2mm}
	TAFT is a plug-in module which enables few-shot segmentation for an existing segmentation network composed of an encoder and a decoder. By transforming the features from the encoder, TAFT provides an effective task-conditioning to the segmentation network. As seen in Figure \ref{figure1}, when an episode with $S$ and $Q$ is given, TAFT computes prototypes $\{\mathbf{c}_{fg},\mathbf{c}_{bg}\}$, which are the pixel averages of the embedded features from the support images for foreground and background classes as done in \cite{PANet,CANet,SG}. 
	Unlike in previous few-shot segmentation methods, however, TAFT also makes use of another set of class references, $\{\mathbf{r}_{fg},\mathbf{r}_{bg}\}$. 
	The reference vectors $\mathbf{r}_{k}$'s are randomly initialized and meta-trained with the encoder $f$ and the decoder $g$ of the segmentation network. 
	While the prototypes $\mathbf{c}_{k}$'s are apparently driven by and thus depend on the current task's input images, $\mathbf{r}_{k}$'s are completely decoupled from the current task. 
	Given the prototypes $\mathbf{c}_{k}$ and references $\mathbf{r}_{k}$, TAFT constructs a linear transformation matrix $\mathbf{P}$ such that the transformed prototype $\mathbf{Pc}_{k}$ is brought near $\mathbf{r}_{k}$ for all $k$.
	
	By using the reference vectors, we can utilize a stable task-agnostic feature space instead of a highly varying feature space from the images of the unseen class. We observe that while the prototypes $\mathbf{c}_{k}$ may change significantly from one task to next as inputs vary, the references $\mathbf{r}_{k}$ change only by little from one episode to next. As such, $\mathbf{r}_{k}$ offers a stable feature space, while the linear transformation $\mathbf{P}$, which is constructed anew in every episode, provides quick task-conditioning. See Figure 5 of Appendix for visual illustration of the stability of the references  $\mathbf{r}_{k}$.
	 The decoder makes use of the query features transformed to the space where $\mathbf{r}_{k}$ reside. In this sense, the stability of  $\mathbf{r}_{k}$ helps the decoder to develop a steady and reliable strategy to generate the segmentation mask.

	Figure \ref{figure1} visualizes 1-way 1-shot segmentation. The reference vectors $\{\mathbf{r}_{fg},\mathbf{r}_{bg}\}$ for foreground and background have the length matching the number of channels in the high-level feature. 
	Encoder $f$ extracts the high-level feature $\mathbf{h} = f(\mathbf{x})$ from the support image $\mathbf{x}$. The downsized labels $\tilde{\mathbf{y}}_{fg}$ and $\tilde{\mathbf{y}}_{bg}$ of the same size as the high-level feature are generated by average pooling. Although we can use a simple resizing when generating the downsized labels, resizing often leads to binary hard labels with all foreground or all background when the object in the image is too large or too small; this is problematic in prototype computation. On the other hand, average pooling generates soft labels with continuous values between 0 and 1, and we can compute both foreground and background prototypes for all cases.   
	Using the downsized soft label $\tilde{\mathbf{y}}_{fg}$ and feature $\mathbf{h}$, the foreground prototype $\mathbf{c}_{fg}$ is computed as
	\vspace{-3.5mm}
	\begin{equation}
		\mathbf{c}_{fg} = \frac{\sum_{i}\sum_{j}\mathbf{h}^{(i,j)}\tilde{\mathbf{y}}_{fg}^{(i,j)}}{\sum_{i}\sum_{j}\tilde{\mathbf{y}}_{fg}^{(i,j)}}
		\label{eq_1}
	\end{equation}
	\vspace{-3.5mm}
	
	where $\mathbf{h}^{(i,j)}$ and $\tilde{\mathbf{y}}_{fg}^{(i,j)}$ denote the pixels in $\mathbf{h}$ and $\tilde{\mathbf{y}}_{fg}$, respectively. Likewise, using soft label $\tilde{\mathbf{y}}_{bg}$ and feature $\mathbf{h}$ the background prototype $\mathbf{c}_{bg}$ is computed as
	
	\vspace{-3.5mm}
	\begin{equation}
		\mathbf{c}_{bg} = \frac{\sum_{i}\sum_{j}\mathbf{h}^{(i,j)}\tilde{\mathbf{y}}_{bg}^{(i,j)}}{\sum_{i}\sum_{j}\tilde{\mathbf{y}}_{bg}^{(i,j)}}
		\label{eq_2}
	\end{equation}
	\vspace{-3.5mm}
	
	where $\tilde{\mathbf{y}}_{bg}^{(i,j)}$ are the pixels in $\tilde{\mathbf{y}}_{bg}$.
	For $N$-shot setting with $N$ support images and support labels, the prototypes $\mathbf{c}_{fg}, \mathbf{c}_{bg}$ are computed as the means of the sample prototypes computed by equations (\ref{eq_1}) and (\ref{eq_2}). Given the prototypes $\{\mathbf{c}_{fg},\mathbf{c}_{bg} \}$ and reference vectors $\{\mathbf{r}_{fg},\mathbf{r}_{bg} \}$, we compute the prototype matrix $\mathbf{C}$ as $\left[\frac{\mathbf{c}_{fg}}{\norm{\mathbf{c}_{fg}}},\frac{\mathbf{c}_{bg}}{\norm{\mathbf{c}_{bg}}}\right]$ and reference matrix $\mathbf{R}$ as $\left[\frac{\mathbf{r}_{fg}}{\norm{\mathbf{r}_{fg}}},\frac{\mathbf{r}_{bg}}{\norm{\mathbf{r}_{bg}}}\right]$. 
	
	We can construct the transformation matrix $\mathbf{P}$ by finding a matrix such that $\mathbf{PC}=\mathbf{R}$. In general, $\mathbf{C}$ is not a square matrix and does not have an inverse. One way to find a reasonable $\mathbf{P}$ is to compute 
	$\mathbf{P} = \mathbf{R}\mathbf{C}^+$, where $\mathbf{C}^+$ is the pseudo-inverse of $\mathbf{C}$ computed as $\{\mathbf{C}^{T}\mathbf{C}\}^{-1}\mathbf{C}^{T}$. This gives the least square fit between $\mathbf{PC}$ and $\mathbf{R}$. 
	Note that in 1-way segmentation the matrix $\mathbf{C}^{T}\mathbf{C}$ is $2 \times 2$, and the inversion of this matrix requires very little computation.
	Matrix $\mathbf{P}$ depends on $\mathbf{C}$, which is driven by the encoder input representing the current task. Thus, it can be said that task-conditioning is achieved via application of the linear transformation using $\mathbf{P}$.
	Given the high-level feature from a query image, TAFT transforms it to the task-agnostic feature pixel-by-pixel using $\mathbf{P}$. 
	The pixel-wise feature transformation can be easily done using a $1 \times 1$ convolution layer with weight $\mathbf{P}$. 
	With this task-adaptive feature transformation, the pixels of the task-agnostic feature get to settle close to the corresponding reference vectors, and the decoder can easily distinguish the foreground and background pixels in the task-agnostic feature. Note that the reference vectors are the only learned part in TAFT. The transformation matrix is not trained but rather computed for each task.
	\vspace{-3mm}
	\subsection{Semantic Enrichment}
	\vspace{-1mm}
	While the proposed TAFT module allows the model to adapt to a new task, the quality of the segmentation prediction is highly dependent on the quality of the high-level features from the encoder. If the encoder fails to provide the informative high-level features, it is impossible to obtain the useful task-agnostic features and therefore the decoder fails to predict the exact segmentation masks. To improve the quality of the high-level features from the encoder, we propose a semantic enrichment (SE) module that utilizes the following two additional components.
	
	First, we adopt a pixel-wise self-attention module on the top of the encoder. Using the self-attention module, we can correlate pixels in the feature with the other feature pixels in the same feature and capture the context among the feature pixels. For the self-attention, we utilize a multi-head scaled dot-product attention proposed in \cite{SA}. To apply the self-attention, we first flatten each feature $X \in \mathbb{R}^{(H \times W \times d)}$ into a one-dimensional feature sequence $X_{seq} \in \mathbb{R}^{(HW \times d)}$. Then, we process the feature sequence through the attention layer and obtain the attended feature sequence having the same size with the input sequence. Finally, we reshape the sequence into a two-dimensional attended feature $X_{attn} \in \mathbb{R}^{(H \times W \times d)}$. By processing the feature using the self-attention module, the high-level features are refined considering the context among the feature pixels, and the quality of semantic information in the high level features is improved.
	
	Our second component in SE module is an auxiliary decoder that conducts multi-class semantic segmentation. While the labels of the 1-way segmentation episodes contain only the information about the target class, there exist multiple objects of various classes in the corresponding training images in general. To make use of the various classes in the training images, we utilize an auxiliary decoder which predicts the segmentation masks of training classes from the features of the training images. For the supervision of the auxiliary decoder, we generate auxiliary segmentation labels for the training images from the original segmentation labels, by making the all test classes as the background. Using the auxiliary decoder and the auxiliary labels, we obtain an auxiliary loss to train the auxiliary decoder and the encoder network. By training the encoder for the various classes in the image via auxiliary loss, the encoder learns to extract the more general features with the richer semantic information which is useful for few-shot segmentation. 
	
	Since the attention module operates on the high-level feature reduced to 1/16 of the original image, the pixel-wise attention does not require significant additional computation. Note that the auxiliary decoder is used only during training and not utilized for evaluation.
	\vspace{-3mm}
	\subsection{Combination with Segmentation Algorithm}
	\vspace{-2mm}
	Algorithm \ref{alg} shows processing of a given episode for the TAFT-SE modules combined with a segmentation algorithm during meta-training (which is the same as the inference process except for the loss computation, the auxiliary decoder and parameter update part). The prototypes are computed using the downsized labels and the pixel-wise attended encoder features from the support set (line 2 to 10). Then the transformation matrix $\mathbf{P}$ is constructed using the prototypes and reference vectors (line 11 to 12). 		
	For training the reference vectors, we utilize a regression loss $L_R$. Using the reference vectors and task-agnostic features, we generate the downsized predictions and compare them with downsized labels (line 16). In prediction, we use the pixel-wise inner product with the reference vectors and normalize the prediction scores for each pixel using the softmax activation function. Since the downsized labels contain continuous values in $[0,1]$, the mean-squared-error (MSE) loss for regression is utilized as the regression loss and computed over all pixels of downsized predictions. Using $L_R$, the reference vectors are trained so that the pixels in task-agnostic features can be easily distinguished. 
	Using the task-agnostic features, the decoder predicts the segmentation masks (line 17). The segmentation loss $L_S$ is computed between the labels and the predicted segmentation masks (line 18). The segmentation loss is a cross-entropy loss computed over all pixels in images, which is generally used for semantic segmentation. Through meta-learning using $L_S$, the decoder learns to generate segmentation masks using the task-agnostic features where each pixel is close to the corresponding reference vector. At the same time, the auxiliary decoder predicts the segmentation masks and the auxiliary segmentation loss is computed using the auxiliary labels. The auxiliary loss is also a cross-entropy loss computed over whole pixels as the segmentation loss (line 19). The auxiliary loss helps the encoder learn to extract more general features. 
	
	\begin{algorithm*}[h]
		\small
		\caption{\small{Procedure for TAFT-SE combined with encoder $f$ and decoder $g$ of a segmentation model to process an episode during meta-training. The episode is composed of $N$ support images/labels and $M$ query images/labels. Label $\mathbf{y}_n$ is composed of foreground label $\mathbf{y}_{n,fg}$ and background label $\mathbf{y}_{n,bg}$. $\mathbf{r}_{fg}$ and $\mathbf{r}_{bg}$ are the reference vectors for foreground and background, respectively. $\mathbf{y}^{(i,j)}$ and $\mathbf{h}^{(i,j)}$ denote the $(i,j)$th pixel of $\mathbf{y}$ and $\mathbf{h}$, respectively. $\mathbf{\check{y}}_n$ denotes the auxiliary label containing supervision for all the training classes.}}
		\label{alg}
		\textbf{Input}: Encoder $f$, decoder $g$, reference vectors $[\mathbf{r}_{fg},\mathbf{r}_{bg}]$, attention module $A$, auxiliary decoder $\check{g}$ \\ $S=\{(\mathbf{x}_{n}, \mathbf{y}_{n})\}_{n=1}^{N}$, $Q=\{(\bar{\mathbf{x}}_{m}, \bar{\mathbf{y}}_{m}, \check{\mathbf{y}}_{m})\}_{m=1}^{M}$
		
		\begin{algorithmic}[1]
			\STATE $L_s\leftarrow0,\;\;\;\; L_r\leftarrow0,\;\;\;\; L_{aux}\leftarrow0 $
			\FOR{$n$ in $ \left \{1 , ... , N \right \}$}
			\FOR{$k$ in $ \left \{{fg,bg} \right \}$}
			\STATE $ \tilde{\mathbf{y}}_{n,k} \leftarrow \text{AvgPool}(\mathbf{y}_{n,k})$\ \ \ \ \ \ \ \ \ \ \ \quad $\triangleright$ \ $ \mathbf{y}_{n,k} \in \mathbb{R}^{(H \times W)},\;\; \tilde{\mathbf{y}}_{n,k} \in \mathbb{R}^{(H_s \times W_s)} $
			
			\STATE $\mathbf{c}_{n,k}\leftarrow \sum_{i}\sum_{j}A(f(\mathbf{x}_n))^{(i,j)}\tilde{\mathbf{y}}_{n,k}^{(i,j)}/\sum_{i}\sum_{j}\tilde{\mathbf{y}}_{n,k}^{(i,j)}$
			\ENDFOR
			\ENDFOR
			\FOR{$k$ in $ \left \{{fg,bg} \right \}$}
			\STATE $\mathbf{c}_{k} \leftarrow \sum_n{\mathbf{c}_{n,k}}/N_s $
			\ENDFOR
			\STATE $\mathbf{C} \leftarrow [\mathbf{c}_{fg}/\norm{\mathbf{c}_{fg}},\mathbf{c}_{bg}/\norm{\mathbf{c}_{bg}}], \;\;\;\; \mathbf{R} \leftarrow [\mathbf{r}_{fg}/\norm{\mathbf{r}_{fg}},\mathbf{r}_{bg}/\norm{\mathbf{r}_{bg}}] $
			
			\STATE $\mathbf{P} \leftarrow \mathbf{R}\{\mathbf{C}^{T}\mathbf{C}\}^{-1}\mathbf{C}^{T}$
			
			\FOR{$m$ in $ \left \{1 , ... , M\right \}$}
			\STATE $\mathbf{h}_a^{(i,j)} \leftarrow \mathbf{P}A(f(\mathbf{\bar{x}}_m))^{(i,j)}$ \ \ \ \ \ \ \ \ \ \ \ \ \ \ \ \ \quad $\triangleright$ $\mathbf{h}_a$ denotes the task-agnostic feature
			\STATE $ \tilde{\mathbf{y}}_{m,fg} \leftarrow \text{AvgPool}(\bar{\mathbf{y}}_{m,fg}), \;\;\;\; \tilde{\mathbf{y}}_{m,bg} \leftarrow \text{AvgPool}(\bar{\mathbf{y}}_{m,bg}) $
			
			\STATE $L_{R} \leftarrow L_{R} + \displaystyle\frac{1}{2\times H_s\times W_s} \displaystyle\sum_{k\in\{{fg,bg}\}}\displaystyle\sum_{i}\sum_{j}\text{MSE}\Bigg\{ \frac{\exp\big(\mathbf{r}_{k}\mathbf{h}_a^{(i,j)}\big)}{\exp\big(\mathbf{r}_{fg}\mathbf{h}_a^{(i,j)}\big)+\exp\big(\mathbf{r}_{bg}\mathbf{h}_a^{(i,j)}\big)}, \tilde{\mathbf{y}}_{n,k}^{(i,j)}\Bigg\}$
			\STATE $\hat{\mathbf{y}}_n \leftarrow g(\mathbf{h}_a)$ 
			\STATE $L_{S} \leftarrow L_{S} + \displaystyle\frac{1}{ H\times W} \sum_{i}\sum_{j}\text{CrossEntropy}(\mathbf{\hat{y}}_{n,fg}^{(i,j)},\bar{\mathbf{y}}_{n}^{(i,j)}) $
			\STATE $L_{aux} \leftarrow L_{aux} + \displaystyle\frac{1}{ H\times W} \sum_{i}\sum_{j}\text{CrossEntropy}({\check{g}(A(f(\mathbf{\bar{x}}_m)))}^{(i,j)},\check{\mathbf{y}}_{n}^{(i,j)}) $
			\ENDFOR
			\STATE Update $f$, $A$ minimizing $L_R+L_S+L_{aux}$; update $g$ minimizing $L_S$; update $[\mathbf{r}_{fg},\mathbf{r}_{bg}]$ minimizing $L_R$; update $\check{g}$ minimizing $L_{aux}$.
		\end{algorithmic}
		\textbf{Output}: Updated encoder $f$, decoder $g$, reference vectors  $[\mathbf{r}_{fg},\mathbf{r}_{bg}]$, attention module $A$, auxiliary decoder $\check{g}$
	\end{algorithm*}
	
	In this paper, we test the TAFT-SE modules in conjunction with Deeplab V3+. Deeplab V3+ consists of an encoder, a decoder, and the Atrous Spatial Pyramid Pooling (ASPP) module. For segmentation, the high-level feature containing the semantic information is extracted from the image by the encoder. The low-level feature including the shape information is also extracted from the middle layer of the encoder. The high-level feature is then processed by the ASPP module to capture the multi-scale information, and the decoder network generates the segmentation mask using the ASPP output feature and the low-level feature.
	
	\begin{figure*}[h]
		\centering
		\includegraphics[width=0.8\textwidth]{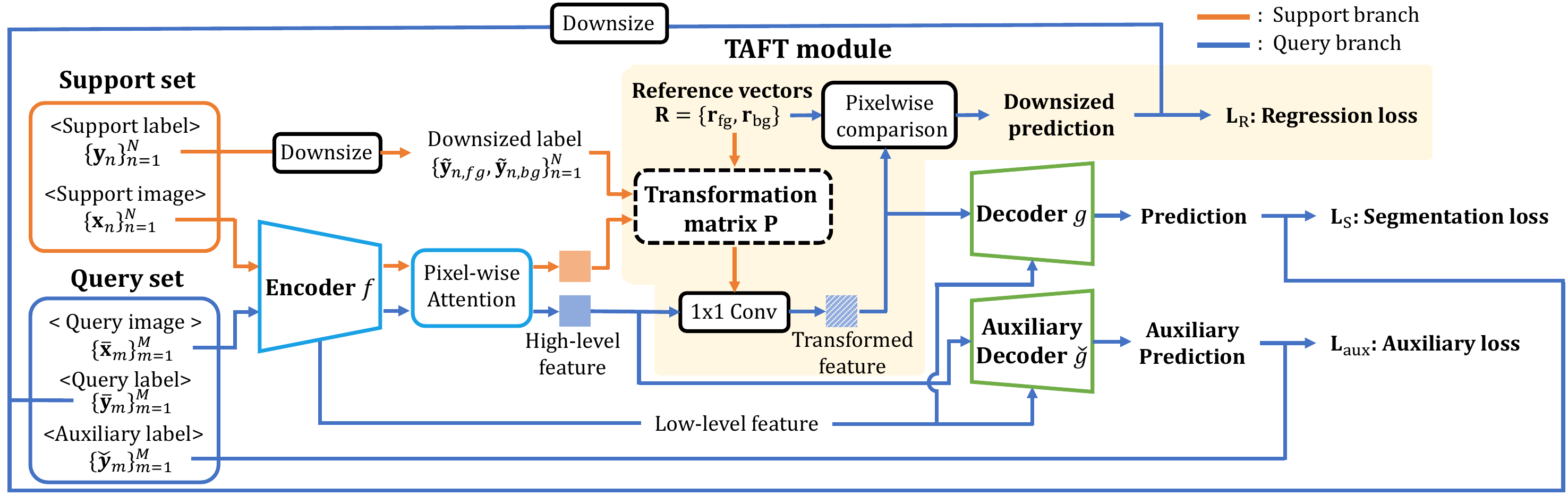}
		\vspace{-2mm}
		\caption{Architecture of the Deeplab V3+ combined with TAFT-SE}
		\label{figure3}
	\end{figure*}

	Figure \ref{figure3} illustrates how TAFT-SE modules are plugged in with Deeplab V3+. The attention module is applied on the top of encoder and refines the high-level features. TAFT operates on the high-level feature and generates the task-agnostic features. The task-agnostic features are then further processed by the ASPP module, which is included as a part of the decoder in the figure. Then, the decoder generates the segmentation masks using the low-level features and the ASPP-processed task-agnostic features together. At the same time, the auxiliary decoder takes high-level features and low-level features as input, and predicts the multi-class segmentation masks.
	Note that in Deeplab V3+, TAFT-SE does not process the low-level feature, since the low-level feature mostly contains the shape information which can be considered general.
	
\begin{figure}[h]
		\centering
		\includegraphics[width=0.4\textwidth]{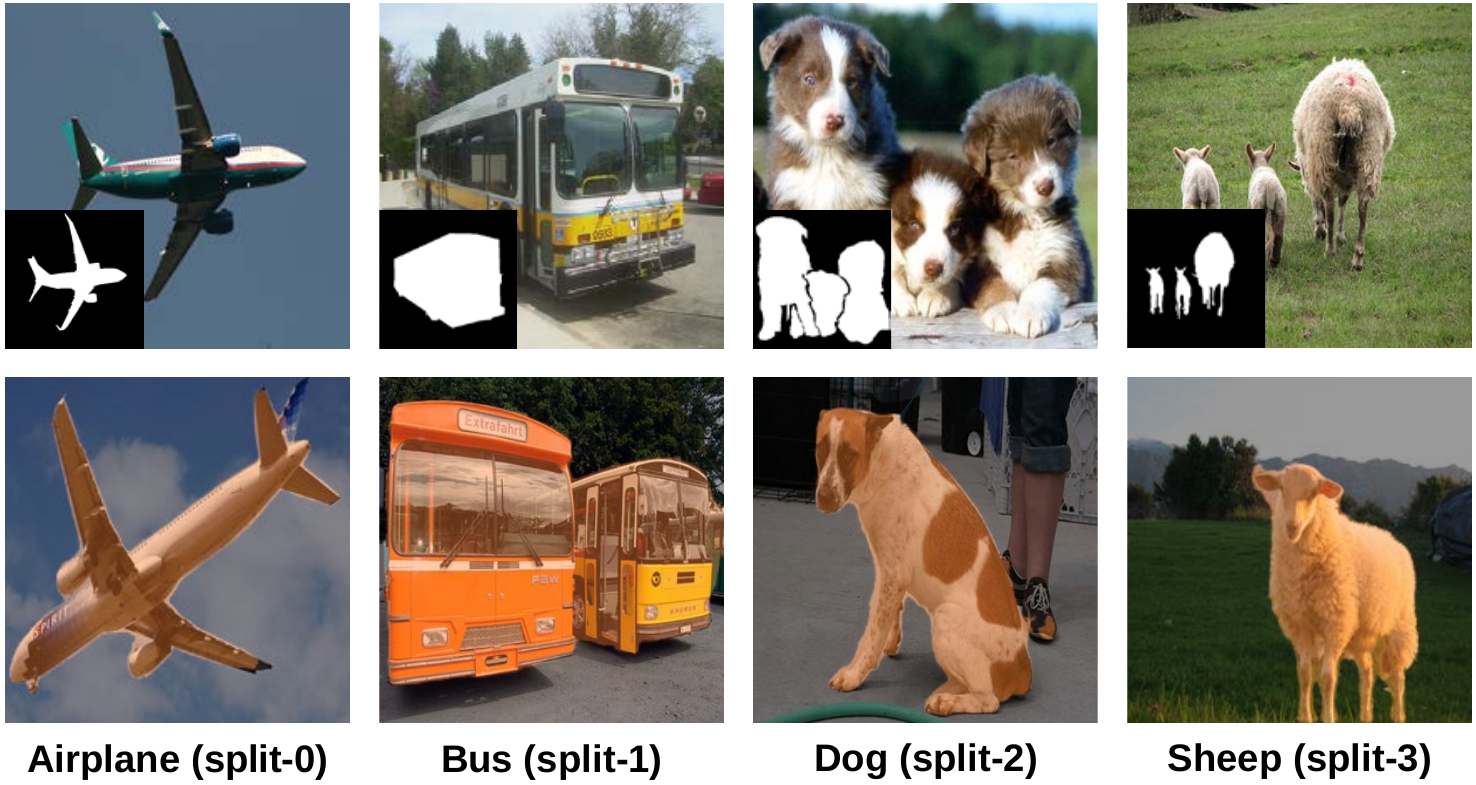}
		\vspace{-2mm}
		\caption{Qualitative results of TAFT-SE on Deeplab V3+ with ResNet-50 backbone for 1-shot segmentation on PASCAL-$5^i$ dataset}
		\label{figure4}
	\end{figure}

	Figure \ref{figure4} displays some qualitative results of TAFT-SE combined with Deeplab V3+. We visualize the 1-shot segmentation results of aeroplane, bus, dog and sheep classes from each split of PASCAL-$5^i$. The images in the first row are support images, with the support labels shown together at the bottom left of the support images. The images in the second and row show the query images with the prediction results. We can see that TAFT-SE on Deeplab V3+ successfully segments the objects from the query images using only a single support sample in each class. More qualitative results can be found in Appendix. 
	
	\vspace{-2mm}
	\section{Related Work}
	\vspace{-1mm}
	\subsection{Semantic Segmentation}
	\vspace{-2mm}
	Semantic segmentation is a computer vision problem that aims to predict the label for each pixel of the image. The fully-convolutional network (FCN) of \cite{FCN} utilizes the convolutional neural network (CNN) architectures designed for classification such as VGG or ResNet by convolutionalization. The convolutionalization process transforms the fully connected layers as $1\times1$ convolution layers. In this way, the network can preserve the location information while predicting the class of each pixel. UNet of \cite{UNet} modifies the FCN architecture to utilize the multi-scale features to obtain better segmentation performance with less data. PSPNet of \cite{PSPNet} proposes a Pyramid Pooling Module (PPM) to effectively aggregate context information over different regions to predict the detailed segmentation masks. Deeplab of \cite{DeeplabV1} and Dilation of \cite{Dilation} introduce a dilated convolution which is able to expand the receptive field without a parameter increase. The advanced versions of Deeplab \cite{DeeplabV3, DeeplabV3+} propose the Atrous Spatial Pyramid Pooling (ASPP) module based on dilated convolution to utilize the multi-scale receptive field. Although the segmentation methods work well when trained with ample data, they cannot be easily adapted to a previously unseen class with only a few data samples, since they are designed to be trained with large datasets. On the other hand, the proposed TAFT-SE modules can be easily combined with an existing segmentation method to enable few-shot segmentation. 
	\vspace{-3mm}
	\subsection{Few-Shot Learning}
	\vspace{-2mm}
	Few-shot learning aims to develop a general classifier that can quickly adapt to the unseen task with only a few labeled samples.
	In carrying out few-shot classification, two main strategies have been developed.
	One is metric-based few-shot learning \cite{TADAM,PN,MN,TapNet}, the goal of which is to learn a mapping to the metric space where samples from the same category cluster together while those from different categories are kept far apart.
	
	Another is the optimization-based meta-learner \cite{MAML,TAML,REPTILE,Ravi,LEO}. This approach aims to train a meta-learner which in turn trains an actual learner. The meta-learner supports learning of the actual learner so that the learner can adapt well to a new task using only a small number of updates based on a few samples. 
	
	Among the few-shot learning algorithms, TapNet of \cite{TapNet} is most closely related to our TAFT module in that
	stand-alone meta-learned reference vectors $\boldsymbol{\Phi}$, which are similar in spirit to $\{\mathbf{r}_{fg},\mathbf{r}_{bg}\}$ of the present method, are employed and linear transformation is used. Despite the similarity, there are some key differences in both philosophical viewpoints as well as design methodologies. In TAFT, the reference vectors are viewed as a more stable form of prototypes residing in a task-agnostic space, which serves as the destination for linear transformation. TAFT brings the embedded features all the way to this task-agnostic space, where inference and reference updates take place. This tends to make the references in TAFT more stable than those in TapNet, helping reliable generation of segmentation masks in the decoder and subsequent training of decoder modules. 
	As for specific design methodology, TapNet's linear transformation attempts to align in-class pairs of vectors while at the same time distancing out-class vectors as much as possible, to maximize classification accuracy. TapNet employs linear nulling of errors for this purpose. In comparison, for TAFT, the least square fit criterion is  more appropriate for linear transformation, as it is well-suited for segmentation.
	
	\vspace{-2mm}
	\subsection{Few-Shot Segmentation}
	\vspace{-2mm}
	OSLSM of \cite{OSLSM} is a first algorithm adopting the few-shot learning strategy to semantic segmentation. OSLSM learns to generate the conditioning parameters for element-wise scaling and shifting. Co-FCN of \cite{coFCN} generates a globally pooled prediction predicts the segmentation mask by fusing it with query features in segmentation branch.
	Recent methods utilize the prototype idea of \cite{PN}, to use the information in the support set efficiently. 
	SG-One of \cite{SG} computes the prototype by the masked-average pooling and compute cosine-similarity map to guide segmentation.
	Following SG-One, many methods utilize the prototype computed by the masked-average pooling for few-shot segmentation \cite {CRNet,PPNet,FWB,PANet,PMM, CANet}. 
	For example, CANet of \cite{CANet} predicts the mask via dense comparison between query feature and prototype, and iteratively updates it using optimization module.
	PANet of \cite{PANet} introduces a novel regularization method of prototype alignment for more effective meta-training.
	PMMs of \cite{PMM}, PPNet of \cite{PPNet}, ASGNet of \cite{ASGNet} and SCL of \cite{SCL} utilize multiple prototypes to leverage the semantic information of different regions of support images.
	Instead of using prototypes, PGNet of \cite{PGNet} and DAN of \cite{DAN} adopt the graph attention mechanism to model the relationship between support feature pixels and query feature pixels. 
	PFENet of \cite{PFENet} utilizes the prior mask computed from pixel-wise cosine similarity between query feature and masked support feature instead of prototype. PFENet also introduces a feature enrichment module which is able to utilize the multi-scale information with hierarchical relations across different scales. CWT of \cite{CWT} utilizes the encoder and decoder pretrained with training classes and adopts a self-attention mechanism to adapt the classifier weights to a given task. MMNet of \cite{MMNet} adopts a learnable memory embeddings to memorize the middle-level meta-class embeddings during meta training and utilize them for few-shot learning.
	
	Our TAFT is similar to the works of \cite{PANet,CANet,SG} in the sense that the prototypes are utilized to represent the pixels. The prior works directly utilize the prototypes that vary widely from task to task in comparing the pixels of the feature with the prototypes. 
	However, the proposed TAFT method does not rely on comparison between the prototypes and the pixels of feature. Instead, TAFT utilizes feature transformation converting task-specific features into fairly stable, task-agnostic features. 
	While the prototype-based approaches require sample-specific comparison between the prototypes and pixels of the feature for each image sample, TAFT does not require any sample-specific process; it generates the transformation matrix $\mathbf{P}$ and use it for every query image. This enables an efficient parallel computation to predict the segmentation masks of different samples at the same time. 
	
	Our SE module is similar to CWT of \cite{CWT} in terms of adopting a self-attention mechanism. While CWT conducts self-attention between features and classifier weights to adapt the model to the given task, our attention module performs self-attention among the pixels from the same feature to enrich the semantic information in the feature. Our SE module and CWT are also similar in terms of conducting semantic segmentation for all training classes during training. However, while CWT utilizes the encoder and decoder pretrained for semantic segmentation task in few-shot adaptation, our SE module conducts the multi-class segmentation only to assist the training of the encoder through the auxiliary loss.
	\vspace{-2mm}
	\section{Experiment Results}
	\vspace{-1mm}
	\label{section4}
	\subsection{Dataset and Evaluation Metric}
	\vspace{-2mm}
	\textbf{PASCAL-}$\mathbf{5^i}$ is used for our experiments. PASCAL-$5^i$ is a dataset based on PASCAL VOC 2012, proposed by \cite{OSLSM} for few-shot segmentation. The 20 classes of VOC 2012 are divided into 4 splits, and each split contains 5 disjoint classes. In our experiments, the 5 classes in one of 4 splits are selected as the test classes, and the remaining classes are used as the training classes. The training samples of training classes and the test samples of test classes are used for training and testing, respectively. 
	The details of class split can be found in \cite{OSLSM}. 
	Data augmentation is not applied.
	
	\textbf{COCO-}$\mathbf{20^i}$ is also utilized for our experiment. It is constructed from the MS-COCO dataset. The 80 classes of MS-COCO are divided into 4 splits, and each split contains 20 disjoint classes. We utilize the class split suggested in \cite{FWB}; the details of the split can be found in \cite{FWB}. Again, data augmentation is not applied.
	
	We use the mean Intersection-over-Union(mIoU) score as the evaluation metric. As done in prior works, we compute the foreground IoU for each class, and use the averaged per-class foreground IoU as mIoU. We also report the evaluation result with another metric, the Foreground-Background IoU(FBIoU) score suggested in \cite{coFCN}. The FBIoU score is computed as the average of foreground IoU and background IoU computed over all test images. 
	
	\vspace{-2mm}
	\subsection{TAFT-SE Architecture}
	\vspace{-2mm}
	TAFT is implemented as a single $1 \times 1$ convolutional layer between encoder and decoder. The weights of this layer reflects the elements of the linear transformation matrix $\textbf{P}$, so that every pixels of feature can be transformed by $\textbf{P}$.
	
	For the self-attention in SE, we adopt a scale-dot product attention layer with a single head. In the attention module, the dimensions of query-key-value in self-attention are the same as the input feature dimension. The auxiliary decoder has the same shape with the original decoder, except the last convolutional layer that conducts the pixel-wise classification; the auxiliary decoder has more output channels for more classes.
	\vspace{-2mm}
	\subsection{Experimental Settings}
	\vspace{-2mm}
	For training TAFT-SE on Deeplab V3+, we use stochastic gradient descent with the learning rate of 0.01 and a momentum of 0.9 for both PASCAL-$5^i$ and COCO-$20^i$ datasets. For the encoder, we utilize ResNet-50 and ResNet-101 networks. We initialize the encoder with the ImageNet pretrained ResNet-50 or ResNet-101, so we apply a 10 times smaller learning rate to the encoder as done in Deeplab V3+. 
	
	We modified the strides in the ResNet encoders so that the low-level features and high-level features are downsized by a factor of 4 and 16, respectively. Since PASCAL-$5^i$ include many small objects, we use the optimized atrous rates of $[1,4,7,11]$ in the ASPP module. 
	
	For all experiments, $3\times10^4$ episodes are used for meta-training and the learning rate is decayed by a factor of 10 after training $2 \times 10^{4}$ episodes. The weight decay with the rate of $3 \times 10^{-4}$ is applied for regularization. Training is done using episodes with 12 queries.
	
	For every experiment, we use the images and labels resized to $512 \times 512$, considering the crop size used in Deeplab V3+. In evaluation, 30,000 episodes are used. For PASCAL-$5^i$ experiments, the multi-scale input test with scale factors $[0.7, 1, 1.3]$ is used as done in prior works of \cite{PGNet,CANet}. In evaluation, the support and query samples are scaled with a specific ratio, and the corresponding predictions are scaled again into the original ratio. The final prediction is done by averaging 3 predictions from different scales. 
	\vspace{-2mm}
	\subsection{Experimental Results}
	\vspace{-2mm}
	\begin{table*}[h]
		\scriptsize
		\centering
		\renewcommand{\arraystretch}{0.95}
		\begin{tabular}{c|c|c|c|c|c|c|c}
			\toprule  
			\multirow{2}{*}{\textbf{Models}} &\multirow{2}{*}{\textbf{Backbone}} &\multicolumn{2}{c|}{mIoU} &\multirow{2}{*}{ $\Delta$} &\multicolumn{2}{c|}{FBIoU} &\multirow{2}{*}{ $\Delta$}\\
			\cmidrule{3-4} \cmidrule{6-7}
			&&1-shot&5-shot & & 1-shot & 5-shot &\\
			\midrule
			\textbf{CANet} \cite{CANet} &\multirow{11}{*}{ResNet-50} & 55.4 &57.1  &1.7 & 66.2 &69.6 &3.4\\
			
			\textbf{PGNet} \cite{PGNet}&  & 56.0 &58.5 &2.5& 69.9 &70.5 &0.6 \\\
			
			\textbf{CRNet} \cite{CRNet}& & 55.7 &58.8  &3.1& 69.9 &71.5  &1.6\\
			\textbf{RPMMs} \cite{PMM} &  & 56.34 &57.30  &0.96 &-&-&- \\
			\textbf{PPNet} \cite{PPNet} & & 51.50 &61.96 &\textbf{10.46}& 69.19 &75.76 &\textbf{6.17}\\
			
			\textbf{PFENet} \cite{PFENet} & & 60.8 &61.9 &1.1 & \textbf{73.3} &73.9  &0.6 \\
			\textbf{ASGNet} \cite{ASGNet} & & 59.29 &63.94 &4.65 & 69.2 &74.2 &5.0 \\
			\textbf{SCL (PFENet)} \cite{SCL} & & \textbf{61.8} &62.9 &1.1 &71.9 &72.8 &0.9 \\
			\textbf{CWT} \cite{CWT} & & 56.4 &63.7 &7.3 &- &- &- \\
			\textbf{MMNet} \cite{MMNet} & & 60.2 &61.8 &1.6&- &- &- \\
			\textbf{TAFT-SE on Deeplab V3+}(Ours)  & &56.69 &\textbf{65.16}  &8.47 &72.51 &\textbf{77.68} & 5.17 \\
			\midrule
			\textbf{FWB} \cite{FWB}   &\multirow{6}{*}{ResNet-101} & 56.2 &59.92 &3.73 &-&-&- \\
			\textbf{DAN} \cite{DAN} &    &58.2&60.5 &2.3  & 71.9 &72.3  &0.4\\
			\textbf{PFENet} \cite{PFENet} & & \textbf{60.1} &61.4 &1.3 & 72.9 &73.5  &0.6 \\
			\textbf{ASGNet} \cite{ASGNet} & & 59.31 &64.36 &5.05 &71.7 &75.2 &3.5 \\
			\textbf{CWT} \cite{CWT} & & 58.0 &64.7 &6.7 &- &- &- \\
			\textbf{TAFT-SE on Deeplab V3+} (Ours)&  &57.98 &\textbf{67.50}  &\textbf{9.52 }&\textbf{73.48} &\textbf{79.40 }&\textbf{5.92} \\
			\bottomrule
		\end{tabular}
		\vspace{-2mm}
		\caption{Mean Intersection-over-Union (mIoU) scores and Foreground-Background Intersection-over-Union (FBIoU) scores averaged over all splits for 1-way PASCAL-$5^i$. $\Delta$ denotes the difference between 1-shot and 5-shot.}
		\label{table1}
	\end{table*}

	\begin{table}[h]
		\centering
		\scriptsize
		\setlength{\tabcolsep}{3.5pt}
		\renewcommand{\arraystretch}{0.95}
		\begin{tabular}{c|c|c|c|c}
			\toprule  
			\multirow{2}{*}{\textbf{Models}}&\multirow{2}{*}{\textbf{Backbone}} &\multicolumn{2}{c|}{mIoU} &\multirow{2}{*}{ $\Delta$} \\
			\cmidrule{3-4}
			&&1-shot&5-shot &\\
			\midrule
			\textbf{PANet} \cite{PANet} &\multirow{7}{*}{ResNet-50} & 20.9 &29.7 &8.8 \\
			\textbf{RPMMs} \cite{PMM}  & &30.58&35.52 &4.94 \\
			\textbf{PPNet} \cite{PPNet} & & 29.03 &38.53 &9.50 \\
			\textbf{ASGNet} \cite{ASGNet} & & 34.56 &42.48 &7.92 \\
			\textbf{CWT} \cite{CWT}&  &32.9 &41.3 &8.4 \\
			\textbf{MMNet} \cite{MMNet}&  &\textbf{37.2} &38.0 &0.8 \\
			\textbf{TAFT-SE on Deeplab V3+}(Ours) & &32.62 &\textbf{45.64}&\textbf{13.62}\\
			\midrule
			\textbf{FWB} \cite{FWB} &\multirow{6}{*}{ResNet-101}  & 21.19 &23.65 &2.46 \\
			\textbf{DAN} \cite{DAN} && 24.4 &29.6 &5.2 \\
			\textbf{PFENet} \cite{PFENet}&  &32.4 &37.4 &5.0 \\
			\textbf{SCL (PFENet)} \cite{SCL} & & \textbf{37.0} &39.9&2.9 \\
			\textbf{CWT} \cite{CWT}&  &32.4 &42.0 &9.6 \\
			\textbf{TAFT-SE on Deeplab V3+} (Ours)  & &33.89 &\textbf{47.01}&\textbf{13.12}\\
			\bottomrule
		\end{tabular}
		\vspace{-2mm}
		\caption{Mean Intersection-over-Union (mIoU) scores averaged over all splits for 1-way COCO-$20^i$.}
		\vspace{-5mm}
		\label{table2}
	\end{table}
	
	\begin{table}[h]
		\scriptsize
		\centering
		\renewcommand{\arraystretch}{0.95}
		\begin{tabular}{c|c|c|c|c|c}
			\toprule  
			TAFT & $Attn$&{$L_{aux}$}&{$MS$}&{1-shot}&{5-shot}\\
			\midrule
			\cmark &\xmark &\xmark&\xmark&51.41&62.14\\
			\cmark &\cmark &\xmark&\xmark&52.80&62.85\\
			\cmark &\cmark &\cmark&\xmark&55.23&64.33\\
			\cmark &\cmark &\cmark&\cmark&\textbf{56.69}&\textbf{65.16}\\
			\bottomrule
		\end{tabular}
		\vspace{-2mm}
		\caption{Results from ablation experiments using PASCAL-$5^i$ dataset. $Attn$ denotes an attention module, $L_{aux}$ denotes an auxiliary loss and the corresponding auxiliary decoder, and $MS$ denotes a multi-scale evaluation. Scores for 1-shot and 5-shot are mIoU scores averaged over all splits.}
		\label{table3}
	\end{table}
	
	We compare TAFT-SE on Deeplab V3+ with prior approaches. In Table \ref{table1}, we display the mean mIoU and FBIoU scores for both 1-shot and 5-shot segmentations for PASCAL-$5^i$. For both ResNet-50 and ResNet-101 backbones, our 1-shot mean mIoU scores fall somewhat lower than previous SOTA methods.
	However, for 5-shots, TAFT-SE modules do indeed extend Deeplab V3+'s few-shot capability to beyond what is possible with the SOTA performance by prior methods by a fair margin. Accordingly, the $\Delta$ value, which represents the ability of the few-shot segmentator to learn from an increasing number of shots, is also the largest except for PPNet in ResNet-50 experiments, and the largest in ResNet-101 experiments. In fact, as shown in Table 7 of Appendix, when compared with PFENet (which gives a SOTA 1-shot mIoU performance in ResNet-101 experiments) across a wider range of 1, 3, 5, 7 and 10 shots, our method already starts to outperform PFENet from 3 shots and on, by a significant margin.
	
	For FBIoU scores, our TAFT-SE on Deeplab V3+ with ResNet-50 backbone falls short of SOTA performance on 1-shot. But for 5-shot, our extension yields the best FBIoU score. In ResNet-101 experiments, TAFT-SE extension exhibits the best 1-shot and 5-shot scores and the best $\Delta$ value, highlighting its ability to learn quickly from an increasing number of examples.  
	The higher FBIoU score means that the proposed method achieves a higher background IoU than prior works. We conjecture that the usage of background pixel information results in higher background IoU and FBIoU scores, as in PANet of \cite{PANet} and PPNet of \cite{PPNet}. While most prior works only utilize the foreground prototype or foreground pixel information, 
	TAFT-SE utilizes the prototypes and reference vectors for both foreground and background. 
	
	Table \ref{table2} displays the mean mIoU scores for COCO-$20^i$. Again, for both ResNet-50 and ResNet-101 backbone, our TAFT-SE extension on Deeplab V3+ gives the best 5-shot performance with large margins.

	Table \ref{table3} shows the results of our ablation experiments with PASCAL-$5^i$ dataset. For the all ablation experiments, we use Deeplab V3+ with the ResNet-50 backbone as a base segmentator, with the same hyperparameter setting. We can see that the attention module and the auxiliary loss consistently improve the performances. For both 1-shot and 5-shot, the auxiliary loss provides more gains compared to the attention module. In Appendix, we show the performance of TAFT-SE extension on FCN of \cite{FCN}, another well-known segmentation model.  
	\vspace{-2mm}
	\section{Conclusion}
	\vspace{-2mm}
	In this paper, we propose a task-adaptive feature transformer module to extend existing segmentation models to
	acquire few-shot segmentation capabilities. TAFT transforms high-level features containing the semantic information into task-agnostic features. 
	The feature transformation converts the semantic information in the high-level features to a form more suitable for the decoder to generate the segmentation mask. 
	We also introduce a semantic enrichment module that enriches the semantic information in the high-level features using the pixel-wise self-attention and the auxiliary segmentation network performing multi-class semantic segmentation.
	The proposed TAFT-SE modules can be easily plugged into existing semantic segmentation algorithms such as Deeplab V3+. Extensive experiments show that while TAFT-SE extensions of existing segmentators are not always SOTA on 1-shot setting, TAFT-SE quickly improves as more shots of examples become available. On 5-shot, TAFT-SE on Deeplab V3+ beats all known tailored few-shot segmentation models.

	\bibliography{egbib}

\begin{thebibliography}{43}
\providecommand{\natexlab}[1]{#1}
\providecommand{\url}[1]{\texttt{#1}}
\expandafter\ifx\csname urlstyle\endcsname\relax
  \providecommand{\doi}[1]{doi: #1}\else
  \providecommand{\doi}{doi: \begingroup \urlstyle{rm}\Url}\fi

\bibitem[Amirreza~Shaban \& Boots(2017)Amirreza~Shaban and Boots]{OSLSM}
Amirreza~Shaban, Shray~Bansal, Z. L. I.~E. and Boots, B.
\newblock One-shot learning for semantic segmentation.
\newblock In \emph{BMVC}, 2017.

\bibitem[Chen et~al.(2017{\natexlab{a}})Chen, Papandreou, Kokkinos, Murphy, and
  Yuille]{DeeplabV1}
Chen, L.-C., Papandreou, G., Kokkinos, I., Murphy, K., and Yuille, A.~L.
\newblock Deeplab: Semantic image segmentation with deep convolutional nets,
  atrous convolution, and fully connected crfs.
\newblock \emph{IEEE transactions on pattern analysis and machine
  intelligence}, 40\penalty0 (4):\penalty0 834--848, 2017{\natexlab{a}}.

\bibitem[Chen et~al.(2017{\natexlab{b}})Chen, Papandreou, Schroff, and
  Adam]{DeeplabV3}
Chen, L.-C., Papandreou, G., Schroff, F., and Adam, H.
\newblock Rethinking atrous convolution for semantic image segmentation.
\newblock \emph{arXiv preprint arXiv:1706.05587}, 2017{\natexlab{b}}.

\bibitem[Chen et~al.(2018)Chen, Zhu, Papandreou, Schroff, and Adam]{DeeplabV3+}
Chen, L.-C., Zhu, Y., Papandreou, G., Schroff, F., and Adam, H.
\newblock Encoder-decoder with atrous separable convolution for semantic image
  segmentation.
\newblock In \emph{Proceedings of the European conference on computer vision
  (ECCV)}, pp.\  801--818, 2018.

\bibitem[Finn et~al.(2017)Finn, Abbeel, and Levine]{MAML}
Finn, C., Abbeel, P., and Levine, S.
\newblock Model-agnostic meta-learning for fast adaptation of deep networks.
\newblock In \emph{International Conference on Machine Learning}, pp.\
  1126--1135, 2017.

\bibitem[Fu et~al.(2019)Fu, Zhang, Zhang, Yan, Chang, Zhang, and Sun]{FSD1}
Fu, K., Zhang, T., Zhang, Y., Yan, M., Chang, Z., Zhang, Z., and Sun, X.
\newblock Meta-ssd: Towards fast adaptation for few-shot object detection with
  meta-learning.
\newblock \emph{IEEE Access}, 7:\penalty0 77597--77606, 2019.

\bibitem[He et~al.(2016)He, Zhang, Ren, and Sun]{Resnet}
He, K., Zhang, X., Ren, S., and Sun, J.
\newblock Deep residual learning for image recognition.
\newblock In \emph{Proceedings of the IEEE conference on computer vision and
  pattern recognition}, pp.\  770--778, 2016.

\bibitem[Jamal \& Qi(2019)Jamal and Qi]{TAML}
Jamal, M.~A. and Qi, G.-J.
\newblock Task agnostic meta-learning for few-shot learning.
\newblock In \emph{Proceedings of the IEEE Conference on Computer Vision and
  Pattern Recognition}, pp.\  11719--11727, 2019.

\bibitem[Kang et~al.(2019)Kang, Liu, Wang, Yu, Feng, and Darrell]{FSD2}
Kang, B., Liu, Z., Wang, X., Yu, F., Feng, J., and Darrell, T.
\newblock Few-shot object detection via feature reweighting.
\newblock In \emph{Proceedings of the IEEE International Conference on Computer
  Vision}, pp.\  8420--8429, 2019.

\bibitem[Karlinsky et~al.(2019)Karlinsky, Shtok, Harary, Schwartz, Aides,
  Feris, Giryes, and Bronstein]{FSD3}
Karlinsky, L., Shtok, J., Harary, S., Schwartz, E., Aides, A., Feris, R.,
  Giryes, R., and Bronstein, A.~M.
\newblock Repmet: Representative-based metric learning for classification and
  few-shot object detection.
\newblock In \emph{Proceedings of the IEEE Conference on Computer Vision and
  Pattern Recognition}, pp.\  5197--5206, 2019.

\bibitem[Krizhevsky et~al.(2012)Krizhevsky, Sutskever, and Hinton]{Alexnet}
Krizhevsky, A., Sutskever, I., and Hinton, G.~E.
\newblock Imagenet classification with deep convolutional neural networks.
\newblock In \emph{Advances in neural information processing systems}, pp.\
  1097--1105, 2012.

\bibitem[Li et~al.(2021)Li, Jampani, Sevilla-Lara, Sun, Kim, and Kim]{ASGNet}
Li, G., Jampani, V., Sevilla-Lara, L., Sun, D., Kim, J., and Kim, J.
\newblock Adaptive prototype learning and allocation for few-shot segmentation.
\newblock In \emph{Proceedings of the IEEE/CVF Conference on Computer Vision
  and Pattern Recognition}, pp.\  8334--8343, 2021.

\bibitem[Liu et~al.(2020{\natexlab{a}})Liu, Zhang, Lin, and Liu]{CRNet}
Liu, W., Zhang, C., Lin, G., and Liu, F.
\newblock Crnet: Cross-reference networks for few-shot segmentation.
\newblock In \emph{Proceedings of the IEEE/CVF Conference on Computer Vision
  and Pattern Recognition}, pp.\  4165--4173, 2020{\natexlab{a}}.

\bibitem[Liu et~al.(2020{\natexlab{b}})Liu, Zhang, Zhang, and He]{PPNet}
Liu, Y., Zhang, X., Zhang, S., and He, X.
\newblock Part-aware prototype network for few-shot semantic segmentation.
\newblock In \emph{European Conference on Computer Vision}, pp.\  142--158.
  Springer, 2020{\natexlab{b}}.

\bibitem[Long et~al.(2015)Long, Shelhamer, and Darrell]{FCN}
Long, J., Shelhamer, E., and Darrell, T.
\newblock Fully convolutional networks for semantic segmentation.
\newblock In \emph{Proceedings of the IEEE conference on computer vision and
  pattern recognition}, pp.\  3431--3440, 2015.

\bibitem[Lu et~al.(2021)Lu, He, Zhu, Zhang, Song, and Xiang]{CWT}
Lu, Z., He, S., Zhu, X., Zhang, L., Song, Y.-Z., and Xiang, T.
\newblock Simpler is better: Few-shot semantic segmentation with classifier
  weight transformer.
\newblock In \emph{Proceedings of the IEEE/CVF International Conference on
  Computer Vision}, pp.\  8741--8750, 2021.

\bibitem[Nguyen \& Todorovic(2019)Nguyen and Todorovic]{FWB}
Nguyen, K. and Todorovic, S.
\newblock Feature weighting and boosting for few-shot segmentation.
\newblock In \emph{Proceedings of the IEEE International Conference on Computer
  Vision}, pp.\  622--631, 2019.

\bibitem[Nichol \& Schulman(2018)Nichol and Schulman]{REPTILE}
Nichol, A. and Schulman, J.
\newblock Reptile: a scalable metalearning algorithm.
\newblock \emph{arXiv preprint arXiv:1803.02999}, 2018.

\bibitem[Oreshkin et~al.(2018)Oreshkin, Lacoste, and Rodriguez]{TADAM}
Oreshkin, B.~N., Lacoste, A., and Rodriguez, P.
\newblock Tadam: Task dependent adaptive metric for improved few-shot learning.
\newblock In \emph{Advances in Neural Information Processing Systems}, 2018.

\bibitem[Rakelly et~al.(2018)Rakelly, Shelhamer, Darrell, Efros, and
  Levine]{coFCN}
Rakelly, K., Shelhamer, E., Darrell, T., Efros, A., and Levine, S.
\newblock Conditional networks for few-shot semantic segmentation.
\newblock \emph{ICLR Workshop}, 2018.

\bibitem[Ravi \& Larochelle(2017)Ravi and Larochelle]{Ravi}
Ravi, S. and Larochelle, H.
\newblock Optimization as a model for few-shot learning.
\newblock In \emph{International Conference on Learning Representations}, 2017.

\bibitem[Redmon et~al.(2016)Redmon, Divvala, Girshick, and Farhadi]{YOLO}
Redmon, J., Divvala, S., Girshick, R., and Farhadi, A.
\newblock You only look once: Unified, real-time object detection.
\newblock In \emph{Proceedings of the IEEE conference on computer vision and
  pattern recognition}, pp.\  779--788, 2016.

\bibitem[Ren et~al.(2015)Ren, He, Girshick, and Sun]{FasterRCNN}
Ren, S., He, K., Girshick, R., and Sun, J.
\newblock Faster r-cnn: Towards real-time object detection with region proposal
  networks.
\newblock In \emph{Advances in neural information processing systems}, pp.\
  91--99, 2015.

\bibitem[Ronneberger et~al.(2015)Ronneberger, Fischer, and Brox]{UNet}
Ronneberger, O., Fischer, P., and Brox, T.
\newblock U-net: Convolutional networks for biomedical image segmentation.
\newblock In \emph{International Conference on Medical image computing and
  computer-assisted intervention}, pp.\  234--241. Springer, 2015.

\bibitem[Rusu et~al.(2018)Rusu, Rao, Sygnowski, Vinyals, Pascanu, Osindero, and
  Hadsell]{LEO}
Rusu, A.~A., Rao, D., Sygnowski, J., Vinyals, O., Pascanu, R., Osindero, S.,
  and Hadsell, R.
\newblock Meta-learning with latent embedding optimization.
\newblock \emph{arXiv preprint arXiv:1807.05960v2}, 2018.

\bibitem[Snell et~al.(2017)Snell, Swersky, and Zemel]{PN}
Snell, J., Swersky, K., and Zemel, R.
\newblock Prototypical networks for few-shot learning.
\newblock In \emph{Advances in Neural Information Processing Systems}, pp.\
  4080--4090, 2017.

\bibitem[Szegedy et~al.(2015)Szegedy, Liu, Jia, Sermanet, Reed, Anguelov,
  Erhan, Vanhoucke, and Rabinovich]{googlenet}
Szegedy, C., Liu, W., Jia, Y., Sermanet, P., Reed, S., Anguelov, D., Erhan, D.,
  Vanhoucke, V., and Rabinovich, A.
\newblock Going deeper with convolutions.
\newblock In \emph{Proceedings of the IEEE conference on computer vision and
  pattern recognition}, pp.\  1--9, 2015.

\bibitem[Szegedy et~al.(2016)Szegedy, Vanhoucke, Ioffe, Shlens, and
  Wojna]{inception}
Szegedy, C., Vanhoucke, V., Ioffe, S., Shlens, J., and Wojna, Z.
\newblock Rethinking the inception architecture for computer vision.
\newblock In \emph{Proceedings of the IEEE conference on computer vision and
  pattern recognition}, pp.\  2818--2826, 2016.

\bibitem[Tian et~al.(2020)Tian, Zhao, Shu, Yang, Li, and Jia]{PFENet}
Tian, Z., Zhao, H., Shu, M., Yang, Z., Li, R., and Jia, J.
\newblock Prior guided feature enrichment network for few-shot segmentation.
\newblock \emph{IEEE Transactions on Pattern Analysis and Machine
  Intelligence}, 2020.

\bibitem[Vaswani et~al.(2017)Vaswani, Shazeer, Parmar, Uszkoreit, Jones, Gomez,
  Kaiser, and Polosukhin]{SA}
Vaswani, A., Shazeer, N., Parmar, N., Uszkoreit, J., Jones, L., Gomez, A.~N.,
  Kaiser, {\L}., and Polosukhin, I.
\newblock Attention is all you need.
\newblock In \emph{Advances in Neural Information Processing Systems}, pp.\
  5998--6008, 2017.

\bibitem[Vinyals et~al.(2016)Vinyals, Blundell, Lillicrap, Kavukcuoglu, and
  Wierstra]{MN}
Vinyals, O., Blundell, C., Lillicrap, T., Kavukcuoglu, K., and Wierstra, D.
\newblock Matching networks for one shot learning.
\newblock In \emph{Advances in Neural Information Processing Systems}, pp.\
  3630--3638, 2016.

\bibitem[Wang et~al.(2020)Wang, Zhang, Hu, Yang, Cao, and Zhen]{DAN}
Wang, H., Zhang, X., Hu, Y., Yang, Y., Cao, X., and Zhen, X.
\newblock Few-shot semantic segmentation with democratic attention networks.
\newblock In \emph{European Conference on Computer Vision}, pp.\  730--746.
  Springer, 2020.

\bibitem[Wang et~al.(2019)Wang, Liew, Zou, Zhou, and Feng]{PANet}
Wang, K., Liew, J.~H., Zou, Y., Zhou, D., and Feng, J.
\newblock Panet: Few-shot image semantic segmentation with prototype alignment.
\newblock In \emph{Proceedings of the IEEE International Conference on Computer
  Vision}, pp.\  9197--9206, 2019.

\bibitem[Wu et~al.(2021)Wu, Shi, Lin, and Cai]{MMNet}
Wu, Z., Shi, X., Lin, G., and Cai, J.
\newblock Learning meta-class memory for few-shot semantic segmentation.
\newblock In \emph{Proceedings of the IEEE/CVF International Conference on
  Computer Vision}, pp.\  517--526, 2021.

\bibitem[Yang et~al.(2020)Yang, Liu, Li, Jiao, and Ye]{PMM}
Yang, B., Liu, C., Li, B., Jiao, J., and Ye, Q.
\newblock Prototype mixture models for few-shot semantic segmentation.
\newblock In \emph{European Conference on Computer Vision}, pp.\  763--778.
  Springer, 2020.

\bibitem[Yoon et~al.(2019)Yoon, Seo, and Moon]{TapNet}
Yoon, S.~W., Seo, J., and Moon, J.
\newblock Tapnet: Neural network augmented with task-adaptive projection for
  few-shot learning.
\newblock In \emph{International Conference on Machine Learning}, pp.\
  7115--7123, 2019.

\bibitem[Yu \& Koltun(2016)Yu and Koltun]{Dilation}
Yu, F. and Koltun, V.
\newblock Multi-scale context aggregation by dilated convolutions.
\newblock \emph{ICLR}, 2016.

\bibitem[Zhang et~al.(2021{\natexlab{a}})Zhang, Xiao, and Qin]{SCL}
Zhang, B., Xiao, J., and Qin, T.
\newblock Self-guided and cross-guided learning for few-shot segmentation.
\newblock In \emph{Proceedings of the IEEE/CVF Conference on Computer Vision
  and Pattern Recognition}, pp.\  8312--8321, 2021{\natexlab{a}}.

\bibitem[Zhang et~al.(2019{\natexlab{a}})Zhang, Lin, Liu, Guo, Wu, and
  Yao]{PGNet}
Zhang, C., Lin, G., Liu, F., Guo, J., Wu, Q., and Yao, R.
\newblock Pyramid graph networks with connection attentions for region-based
  one-shot semantic segmentation.
\newblock In \emph{Proceedings of the IEEE International Conference on Computer
  Vision}, pp.\  9587--9595, 2019{\natexlab{a}}.

\bibitem[Zhang et~al.(2019{\natexlab{b}})Zhang, Lin, Liu, Yao, and Shen]{CANet}
Zhang, C., Lin, G., Liu, F., Yao, R., and Shen, C.
\newblock Canet: Class-agnostic segmentation networks with iterative refinement
  and attentive few-shot learning.
\newblock In \emph{Proceedings of the IEEE Conference on Computer Vision and
  Pattern Recognition}, pp.\  5217--5226, 2019{\natexlab{b}}.

\bibitem[Zhang et~al.(2021{\natexlab{b}})Zhang, Kang, Wei, and Yang]{CyCTR}
Zhang, G., Kang, G., Wei, Y., and Yang, Y.
\newblock Few-shot segmentation via cycle-consistent transformer.
\newblock In \emph{Advances in Neural Information Processing Systems},
  2021{\natexlab{b}}.

\bibitem[Zhang et~al.(2020)Zhang, Wei, Yang, and Huang]{SG}
Zhang, X., Wei, Y., Yang, Y., and Huang, T.~S.
\newblock Sg-one: Similarity guidance network for one-shot semantic
  segmentation.
\newblock \emph{IEEE Transactions on Cybernetics}, 2020.

\bibitem[Zhao et~al.(2017)Zhao, Shi, Qi, Wang, and Jia]{PSPNet}
Zhao, H., Shi, J., Qi, X., Wang, X., and Jia, J.
\newblock Pyramid scene parsing network.
\newblock In \emph{Proceedings of the IEEE conference on computer vision and
  pattern recognition}, pp.\  2881--2890, 2017.

\end{thebibliography}
	\bibliographystyle{icml2022}

	\newpage
	\appendix
	\onecolumn
	\section{Additional Experimental Results }
	In Tables \ref{table_app_1}, \ref{table_app_2} and \ref{table_app_3}, we display mIoU and FBIoU for PASCAL-$5^i$, and mIoU for COCO-$20^i$, for all splits. We show the results of the prior works using the smaller VGG-16 backbone, as well as the results of a concurrent work, Cycle-Consistent Transformer (CyCTR). We also present the performances of TAFT-SE combined with Fully Convolutional Network (FCN).
	\subsection{Combination with FCN}
	In the main paper, we argue that TAFT-SE can be combined with the existing semantic segmentation algorithm achieving few-shot segmentation capability. As an example, we showed that TAFT-SE combined with Deeplab V3+ achieved state-of-the-art few-shot segmentation performance. To further demonstrate the extendibility and versatility of TAFT-SE,  we plug TAFT-SE into another well-known segmentation algorithm, Fully Convolutional Network (FCN).
	FCN utilizes the convolutional neural network (CNN) architectures designed for classification by convolutionalization. The convolutionalization process transforms the fully connected layers as $1 \times 1$ convolution layers. In this way, the network can preserve the location information while predicting the class of each pixel. 
	After prediction, FCN upsamples the scores into the original image size. In FCN, the convolution layers or the feature extractor of classification network is considered as encoder, and the convolutionalized layers and upsampling layers are considered as decoder.  
	
	\begin{figure*}[h]
		\centering
		\includegraphics[width=0.90\textwidth]{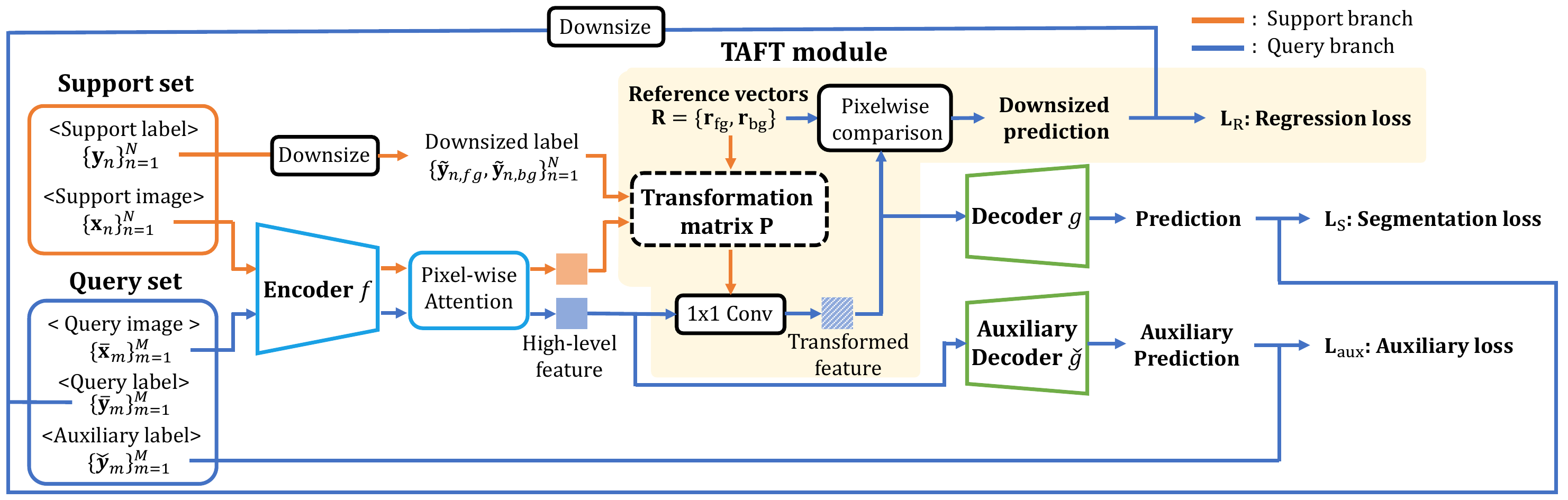}
		\caption{Architecture of the FCN combined with TAFT-SE}
		\label{figure_app_1}
	\end{figure*}
	
	Since FCN can utilize different classification networks, there can be many different variations. Among them, we utilize the FCN architecture with the ResNet-50 and the ResNet-101 encoders provided by the pytorch framework in our experiments. Figure \ref{figure_app_1} illustrates this version of FCN combined with the TAFT-SE modules. 
	The feature transformation by TAFT is applied to the attended high-level feature from the encoder and attention module. The downsized prediction is generated with the task-agnostic feature and the reference vectors, and the regression loss is computed with downsized label and prediction.
	The convolutional decoder generates the segmentation mask with the task-specific high-level feature. The segmentation loss is then computed using the original label and the generated segmentation mask. At the same time, the auxiliary prediction is made by the auxiliary decoder, and the auxiliary segmentation loss is computed from the auxiliary prediction and the auxiliary label.
	For training TAFT-SE on FCN, we utilize the learning rate of $3\times10^{-3}$. We employ the training episodes with 7 queries for 1-shot, and 3 queries for 5-shot. For the remaining hyperparameter, we adopt the same  hyperparameters as those of TAFT-SE on Deeplab V3+.
	In Tables \ref{table_app_1} and \ref{table_app_2}, we display the mean Intersection-over-Union (mIoU) scores and foreground-background Intersection-over-Union (FBIoU) scores of TAFT-SE on FCN and TAFT-SE on Deeplab V3+ for the all splits of PASCAL-$5^i$, in comparison with those of other methods. 
	We can see that for 5-shot, TAFT-SE on FCN shows the mIoU and FBIoU scores comparable to most of the prior works. For FBIoU scores, TAFT-SE on FCN outperforms all the prior works except PPNet in ResNet-50, and performs better than all the other method in ResNet-101. In Table \ref{table_app_3}, we show the mean Intersection-over-Union (mIoU) scores of TAFT-SE on FCN and TAFT-SE on Deeplab V3+ for the all splits of COCO-$20^i$, with the scores of other methods. Again, TAFT-SE on FCN shows the competitive performances compared to SOTA methods. Especially in ResNet-101 experiments, TAFT-SE on FCN outperforms all the prior methods. For $\Delta$ value, TAFT-SE on FCN shows largest $\Delta$ except TAFT-SE on Deeplab V3+.

	\begin{table*}[h]
		\scriptsize
		\centering
		\setlength{\tabcolsep}{5pt}
		\renewcommand{\arraystretch}{0.95}
		\begin{tabular}{c|c|cc|cc|cc|cc|cc|c}
			\toprule  
			\multirow{3}{*}{\textbf{Models}} &\multirow{3}{*}{\textbf{Backbone}} &\multicolumn{10}{c|}{mIoU} &\multirow{3}{*}{ $\Delta$} \\
			\cmidrule{3-12} 
			&& \multicolumn{2}{c|}{split-0} & \multicolumn{2}{c|}{split-1}	& \multicolumn{2}{c|}{split-2} &\multicolumn{2}{c|}{split-3} &\multicolumn{2}{c|}{mean} & \\
			\cmidrule{3-12}
			&& 1-shot&5-shot & 1-shot&5-shot	& 1-shot&5-shot &1-shot&5-shot &1-shot&5-shot & \\
			\midrule
			\textbf{OSLSM} \cite{OSLSM}  &\multirow{5}{*}{VGG-16}  & 33.6 &35.9	&55.3  &58.1& 40.9 &42.7  & 33.5 &39.1 & 40.8 &43.9 &3.1 \\
			\textbf{co-FCN} \cite{coFCN} && 36.7 &37.5	&50.6 &50.0 & 44.9 &44.1  &32.4 &33.9 & 41.1  &41.4&0.3\\
			\textbf{SG-One} \cite{SG}   && 40.2 &41.9	&58.4 &58.6 & 48.4 &48.6  & 38.4 &39.4& 46.3 &47.1 &0.8  \\
			\textbf{PANet} \cite{PANet} &  & 42.3 &\textbf{51.8}	&58.0 &\textbf{64.6} & 51.1  &\textbf{59.8}& 41.2 &46.5 & 48.1 &\textbf{55.7}  &\textbf{7.6}\\
			\textbf{FWB} \cite{FWB}   & & \textbf{47.04}&50.87&\textbf{59.64}&62.86&\textbf{52.61}&56.48&\textbf{48.27}&\textbf{50.09}&\textbf{51.90}&55.08&3.18 \\
			\midrule
			\textbf{CANet} \cite{CANet} &\multirow{13}{*}{ResNet-50} & 52.5 &55.5	&65.9 &67.8 &51.3  &51.9& 51.9 &53.2 & 55.4 &57.1  &1.7\\
			
			\textbf{PGNet} \cite{PGNet}&   & 56.0 &57.9 &66.9  &68.7& 50.6  &52.9& 50.4 &54.6 & 56.0 &58.5 &2.5 \\
			
			\textbf{CRNet} \cite{CRNet}&  & - &-	&- &-& - &-  & - &- & 55.7 &58.8  &3.1 \\
			\textbf{RPMMs} \cite{PMM} &   & 55.16 &56.28	&66.91 &67.34& 52.61 &54.52  & 50.68 &51.00 & 56.34 &57.30  &0.96  \\
			\textbf{PPNet} \cite{PPNet} & & 47.83 &58.39	&58.75 &67.83& 53.80 &64.88 & 45.63 &56.73 & 51.50 &61.96 &10.46\\
			
			\textbf{PFENet} \cite{PFENet} &  & 61.7 &63.1	&69.5 &70.7& 55.4 &55.8  & 56.3 &57.9 & 60.8 &61.9 &1.1 \\
			\textbf{ASGNet} \cite{ASGNet} & & 58.84&63.66&67.86&70.55&56.79&64.17&53.66&57.38&59.29&63.94&4.65 \\
			\textbf{SCL (PFENet)} \cite{SCL} && 63.0&64.5&70.0&70.9&56.5&57.3&57.7&58.7&61.8&62.9&1.1\\
			\textbf{CWT} \cite{CWT} & & 56.3&61.3&62.0&68.5&\textbf{59.9}&\textbf{68.5}&47.2&56.6&56.4&63.7&7.3 \\
			\textbf{MMNet} \cite{MMNet} & &58.0&60.0&70.0&70.6&58.0&56.3&55.0&60.3& 60.2 &61.8 &1.6 \\
			\textbf{CyCTR} \cite{CyCTR} & &\textbf{67.8}&\textbf{71.1}&\textbf{72.8}&\textbf{73.2}&58.0&60.5&\textbf{58.0}&57.5&\textbf{64.2}&\textbf{65.6}&1.4 \\
			\textbf{TAFT-SE on FCN}(Ours)  &&46.75&55.21&60.55&68.79&55.14&64.94&51.12&59.26 &53.39&62.05 &8.66  \\
			\textbf{TAFT-SE on Deeplab V3+}(Ours)  &&49.97&58.27&64.08&70.63&57.11&68.24&55.61&\textbf{63.50} &56.69 &65.16 &8.47  \\
			\midrule
			\textbf{FWB} \cite{FWB}   &\multirow{8}{*}{ResNet-101} & 51.30 &54.84	&64.49&67.38& 56.71 &62.16  & 52.24 &55.30 & 56.19 &59.92 &3.73 \\
			\textbf{DAN} \cite{DAN} &   & 54.7 &57.9	&68.6 &69.0& 57.8 &60.1  &51.6 &54.9 &58.2&60.5 &2.3 \\
			\textbf{PFENet} \cite{PFENet} &&60.5&62.8&69.4&70.4&54.4&54.9&55.9&57.6 & 60.1 &61.4 &1.3 \\
			\textbf{ASGNet} \cite{ASGNet} & &59.84&64.55&67.43&71.32&55.59&64.24&54.39&57.33&59.31&64.36&5.05 \\
			\textbf{CWT} \cite{CWT} & & 56.9&62.6&65.2&70.2&\textbf{61.2}&68.8&48.8&57.2&58.0&64.7&6.7 \\
			\textbf{CyCTR} \cite{CyCTR} & & \textbf{69.3}&\textbf{73.5}&\textbf{72.7}&\textbf{74.0}&56.5&58.6&\textbf{58.6}&60.2&\textbf{64.3}&66.6&2.3 \\
			\textbf{TAFT-SE on FCN}(Ours)  &&49.98&58.57&61.36&69.90&55.15&66.18&53.30&60.68 &54.95&63.83 &8.88  \\
			\textbf{TAFT-SE on Deeplab V3+} (Ours)&&51.70&61.58&64.99&72.60&58.29&\textbf{70.96}&56.92&\textbf{64.86}  &57.98 &\textbf{67.50}  &\textbf{9.52 }\\
			\bottomrule
		\end{tabular}
		\caption{Mean Intersection-over-Union (mIoU) scores for 1-way PASCAL-$5^i$. $\Delta$ denotes the difference between 1-shot and 5-shot.}
		\label{table_app_1}
	\end{table*}

	\begin{table*}[h]
		\scriptsize
		\centering
		\setlength{\tabcolsep}{5pt}
		\renewcommand{\arraystretch}{0.95}
		\begin{tabular}{c|c|cc|cc|cc|cc|cc|c}
			\toprule  
			\multirow{3}{*}{\textbf{Models}} &\multirow{3}{*}{\textbf{Backbone}} &\multicolumn{10}{c|}{mIoU} &\multirow{3}{*}{ $\Delta$} \\
			\cmidrule{3-12} 
			&& \multicolumn{2}{c|}{split-0} & \multicolumn{2}{c|}{split-1}	& \multicolumn{2}{c|}{split-2} &\multicolumn{2}{c|}{split-3} &\multicolumn{2}{c|}{mean} & \\
			\cmidrule{3-12}
			&& 1-shot&5-shot & 1-shot&5-shot	& 1-shot&5-shot &1-shot&5-shot &1-shot&5-shot & \\
			\midrule
			\textbf{OSLSM} \cite{OSLSM}  &\multirow{4}{*}{VGG-16}   & -	&- & -  & -& -  & -& -  & - & 61.3 &61.5&0.2 \\
			\textbf{co-FCN} \cite{coFCN} && -	&- & -  & -& -  & -& -  & - & 60.1  &60.2&0.1\\
			\textbf{SG-One} \cite{SG}   && -	&- & -  & -& -  & -& -  & -& 63.1 &65.9 &2.8  \\
			\textbf{PANet} \cite{PANet} &   & -	&- & -  & -& -  & -& -  & - & 66.5 &70.7 &4.2 \\
			\midrule
			\textbf{CANet} \cite{CANet} &\multirow{9}{*}{ResNet-50} & \textbf{71.0}	&\textbf{74.2 }& 76.7 & 80.3&  54.0& 57.0& 67.2  & 66.8 & 66.2 &69.6 &3.4\\
			
			\textbf{PGNet} \cite{PGNet}&    & -	&- & -  & -& -  & -& -  & -&69.9 &70.5&0.6 \\
			
			\textbf{CRNet} \cite{CRNet}&  & - &-	&- &-& - &-  & - &- & 69.9 &71.5&1.6    \\
			\textbf{PPNet} \cite{PPNet} & & -	&- & -  & -& -  & -& -  & - & 69.19 &75.76 &\textbf{6.17}\\
			\textbf{PFENet} \cite{PFENet} &   & - &-	&- &-& - &-  & - &- & \textbf{73.3} &73.9  &0.6 \\
			\textbf{ASGNet} \cite{ASGNet} & & -	&- & -  & -& -  & -& -  & - &69.2&74.2&5.0 \\
			\textbf{SCL (PFENet)} \cite{SCL} && -	&- & -  & -& -  & -& -  & - &71.9&72.8&0.9\\
			\textbf{TAFT-SE on FCN}(Ours)  &&67.40&71.52&74.44&79.96&68.15&76.07&68.23&74.87 &69.56&75.61 &6.05  \\
			\textbf{TAFT-SE on Deeplab V3+}(Ours)  &&69.04&73.37&\textbf{78.56}&\textbf{81.52}&\textbf{70.27}&\textbf{78.27}&\textbf{72.15}&\textbf{77.55} &72.51 &\textbf{77.68} &5.17  \\
			\midrule
			\textbf{DAN} \cite{DAN} &\multirow{6}{*}{ResNet-101}    & - &-	&- &-& - &-  & - &- & 71.9 &72.3  &0.4 \\
			\textbf{PFENet} \cite{PFENet} & & - &-	&- &-& - &-  & - &- & 72.9&73.5  &0.6 \\
			\textbf{ASGNet} \cite{ASGNet} & & -	&- & -  & -& -  & -& -  & - &71.7&75.2&3.5 \\
			\textbf{CyCTR} \cite{CyCTR} & & -	&- & -  & -& -  & -& -  & -&72.9&75.0&2.1\\
			\textbf{TAFT-SE on FCN}(Ours)  &&68.83&73.71&75.89&80.41&67.38&77.04&69.80&76.30 &70.41&76.87 &\textbf{6.46} \\
			\textbf{TAFT-SE on Deeplab V3+} (Ours)&&\textbf{70.20}&\textbf{75.51}&\textbf{78.82}&\textbf{83.28}&\textbf{72.32}&\textbf{80.39}&\textbf{72.57}&\textbf{78.42}&\textbf{73.48} &\textbf{79.40} &\textbf{5.92}\\
			\bottomrule
		\end{tabular}
		\caption{Foreground-Background Intersection-over-Union (FBIoU) scores for 1-way PASCAL-$5^i$. $\Delta$ denotes the difference between 1-shot and 5-shot.}
		\label{table_app_2}
	\end{table*}

	\begin{table*}[h]
		\scriptsize
		\centering
		\setlength{\tabcolsep}{5pt}
		\renewcommand{\arraystretch}{0.95}
		\begin{tabular}{c|c|cc|cc|cc|cc|cc|c}
			\toprule  
			\multirow{3}{*}{\textbf{Models}} &\multirow{3}{*}{\textbf{Backbone}} &\multicolumn{10}{c|}{mIoU} &\multirow{3}{*}{ $\Delta$} \\
			\cmidrule{3-12} 
			&& \multicolumn{2}{c|}{split-0} & \multicolumn{2}{c|}{split-1}	& \multicolumn{2}{c|}{split-2} &\multicolumn{2}{c|}{split-3} &\multicolumn{2}{c|}{mean} & \\
			\cmidrule{3-12}
			&& 1-shot&5-shot & 1-shot&5-shot	& 1-shot&5-shot &1-shot&5-shot &1-shot&5-shot & \\
			\midrule
			\textbf{PANet} \cite{PANet} &\multirow{9}{*}{ResNet50} & - &-	&- &- &-  &-& - &- &20.9 &29.7  &8.8\\
			\textbf{RPMMs} \cite{PMM} &  & 29.53 &33.82	&36.82&41.96& 28.94&32.99  & 27.02&33.33 &30.58&35.52 &4.94   \\
			\textbf{PPNet} \cite{PPNet} & & 28.09&38.97	&30.81 &40.81& 29.49 &37.07  & 27.70 &37.28 & 29.03 &38.53 &9.50\\
			\textbf{ASGNet} \cite{ASGNet} & & 34.89&40.99&36.94&48.28&34.33&40.10&32.08&40.54&34.56&42.48&7.92\\
			\textbf{CWT} \cite{CWT} & & 32.2&40.1&36.0&43.8&31.6&39.0&31.6&42.4&32.9&41.3&8.4 \\
			\textbf{MMNet} \cite{MMNet} & &34.9&38.5&41.0&39.6&37.8&38.4&35.2&35.5&37.2 &38.0 &0.8 \\
			\textbf{CyCTR} \cite{CyCTR} & & \textbf{38.9}&41.1&\textbf{43.0}&48.9&\textbf{39.6}&45.2&\textbf{39.8}&\textbf{47.0}&\textbf{40.3}&45.6&5.3\\
			\textbf{TAFT-SE on FCN}(Ours)  &&29.96&39.73&32.64&45.25&31.08&41.96&29.64&39.03 &30.83&41.49&10.66  \\
			\textbf{TAFT-SE on Deeplab V3+}(Ours)  &&31.12&\textbf{43.45}&35.23&\textbf{49.26}&33.04&\textbf{46.66}&31.07&43.21 &32.62 &\textbf{45.64} &\textbf{13.02}  \\
			\midrule
			\textbf{FWB} \cite{FWB}   &\multirow{8}{*}{ResNet101}& 16.98 &19.13	&17.98&21.46& 20.96 &23.93  &28.85 &30.08 & 21.19 &23.65 &2.46 \\
			\textbf{DAN} \cite{DAN} &  & - &-	&- &-&- &-  & - &- & 24.4 &29.6 &5.2 \\
			\textbf{PFENet} \cite{PFENet} & & 34.3 &38.5	&33.0 &38.6& 32.3 &38.2 & 30.1 &34.3 & 32.4&37.4 &5.0\\
			\textbf{SCL (PFENet)} \cite{SCL} & &\textbf{36.4}&38.9&\textbf{38.6}&40.5&\textbf{37.5}&41.5&\textbf{35.4}&38.7&\textbf{37.0}&39.9&2.9 \\
			\textbf{CWT} \cite{CWT} & & 30.3&38.5&36.6&46.7&30.5&39.4&32.2&43.2&32.4&42.0&9.6 \\
			\textbf{TAFT-SE on FCN}(Ours)  &&31.00&41.95&32.81&46.00&32.32&44.02&30.37&40.53 &31.63&43.13 &11.50  \\
			\textbf{TAFT-SE on Deeplab V3+} (Ours)&&32.08&\textbf{46.01}&36.10&\textbf{49.61}&34.96&\textbf{48.47}&32.40&\textbf{43.94} &33.89&\textbf{47.01}&\textbf{13.12}\\
			\bottomrule
		\end{tabular}
		\caption{Mean Intersection-over-Union (mIoU) scores for 1-way COCO-$20^i$. $\Delta$ denotes the difference between 1-shot and 5-shot.}
		\label{table_app_3}
	\end{table*}
	\subsection{Comparison with Concurrent Work}	
	As of writing of our present submission, we came across a concurrent work that is worthy of mentioning. CyCTR of \cite{CyCTR} proposes to use a transformer block to establish cycle consistency relationship among query and support features through cross-attention between features. The method is used in conjunction with the existing PFENet architecture.
	Similar to our work, CyCTR transforms the feature to conduct few-shot segmentation. Also, our SE module and CyCTR both utilize the same multi-head scaled-dot product attention for self-attention.  
	Our method, however, is different from CyCTR in the sense that our method transforms the feature via linear transformation, while CyCTR transforms the features nonlinearly through the attention layer. Also, while CyCTR conducts the attention between the pixels from the different features to correlate different features, the attention module in our SE module conducts the self-attention among the pixels from the same feature to capture the context in the feature. Furthermore, CyCTR is applied upon the different few-shot segmentation method of PFENet and enhances few-shot segmentation capability of PFENet. On the other hand, our TAFT-SE method is combined with the semantic segmentation model such as Deeplab V3+, which is not designed for few-shot setting, and newly adds the few-shot segmentation capability to the model.
	CyCTR outperforms the previous SOTA methods in both 1-shot mIoU and 5-shot mIoU, for both PASCAL-$5^i$ and COCO-$20^i$. We compare the performance of CyCTR with our TAFT-SE on Deeplab V3+ in Tables \ref{table_app_1},\ref{table_app_2} and \ref{table_app_3}. For 1-shot mIoU, CyCTR performs better than our TAFT-SE on Deeplab V3+ in both PASCAL-$5^i$ and COCO-$20^i$ datasets, for both ResNet-50 and ResNet-101 backbones. However for 5-shot mIoU, TAFT-SE on Deeplab V3+ outperforms CyCTR in COCO-$20^i$ dataset. For PASCAL-$5^i$, TAFT-SE on Deeplab V3+ shows the similar level of performance in ResNet-50 backbone, and shows better performance in ResNet-101 backbone compared to CyCTR. For PASCAL-$5^i$ FBIoU, our TAFT-SE surpasses the CyCTR in both 1-shot and 5-shot. In all experiments, TAFT-SE on Deeplab V3+ shows greater $\Delta$ than CyCTR.
	\subsection{Additional Experiments with Various Numbers of Shots}
	In order to confirm the quickly improving performance of the TAFT-SE extended segmentator with the number of shots, we  compare mean mIoU scores from  1-shot to 10-shot in Table \ref{table_app_4} between TAFT-SE on Deeplab V3+ and the tailor-designed few-shot segmentator PFENet. Between PASCAL-$5^i$ and COCO-$20^i$, we chose the PASCAL-$5^i$ dataset on which our method showed relatively weaker performance. The results clearly show that PFENet gains only little as the shot number increases. In fact, its 10-shot performance is actually worse than 7-shot performance. On the other hand, TAFT-SE on Deeplab V3+ improves quickly as the number of shots increases, starting to outperform PFENet on 3 shots. The performance gain of the TAFT-SE based extension grows as it is exposed to more shots. 
	\begin{table}[h]
		\centering
		\scriptsize
		\begin{threeparttable}
			\begin{tabular}{c|cc|cc|cc|cc|cc}
				\toprule  
				\multirow{2}{*}{\textbf{Models}}    & \multicolumn{2}{c|}{1-shot} & \multicolumn{2}{c|}{3-shot}	& \multicolumn{2}{c|}{5-shot} &\multicolumn{2}{c|}{7-shot} &\multicolumn{2}{c}{10-shot} \\
				\cmidrule{2-11}
				& mIoU & $\Delta$ & mIoU & $\Delta$	& mIoU & $\Delta$ &mIoU & $\Delta$ &mIoU & $\Delta$ \\
				\midrule
				
				\textbf{PFENet}$^*$ \cite{PFENet}  &\textbf{60.84} &-	&61.60 &0.76& 61.81 &0.21 & 62.03 &0.22 & 61.77 &\textcolor{red}{\textbf{-0.26}}  \\
				
				\textbf{TAFT-SE on Deeplab V3+} &56.69 &-	&\textbf{63.90} &\textbf{7.21}&\textbf{65.16} &\textbf{1.26} &\textbf{65.98}&\textbf{0.82 }&\textbf{66.64} &\textbf{0.66 }\\
				\bottomrule
			\end{tabular}
			\begin{tablenotes}   
				\item $^*$ PFENet performances are measured with our reproduction based on github code of \cite{PFENet}.
			\end{tablenotes}
		\end{threeparttable}
		\caption{1-shot to 10-shot mIoU scores averaged over all splits for 1-way PASCAL-$5^i$. $\Delta$ denotes the score difference between the shot below and the current shot.}
		\label{table_app_4}
		
	\end{table}

	\subsection{Visualization of Prototype and Reference Vectors}	
	
\begin{figure*}[h]
		\centering
		\begin{subfigure}[b]{0.48\textwidth}
			\includegraphics[width=\textwidth]{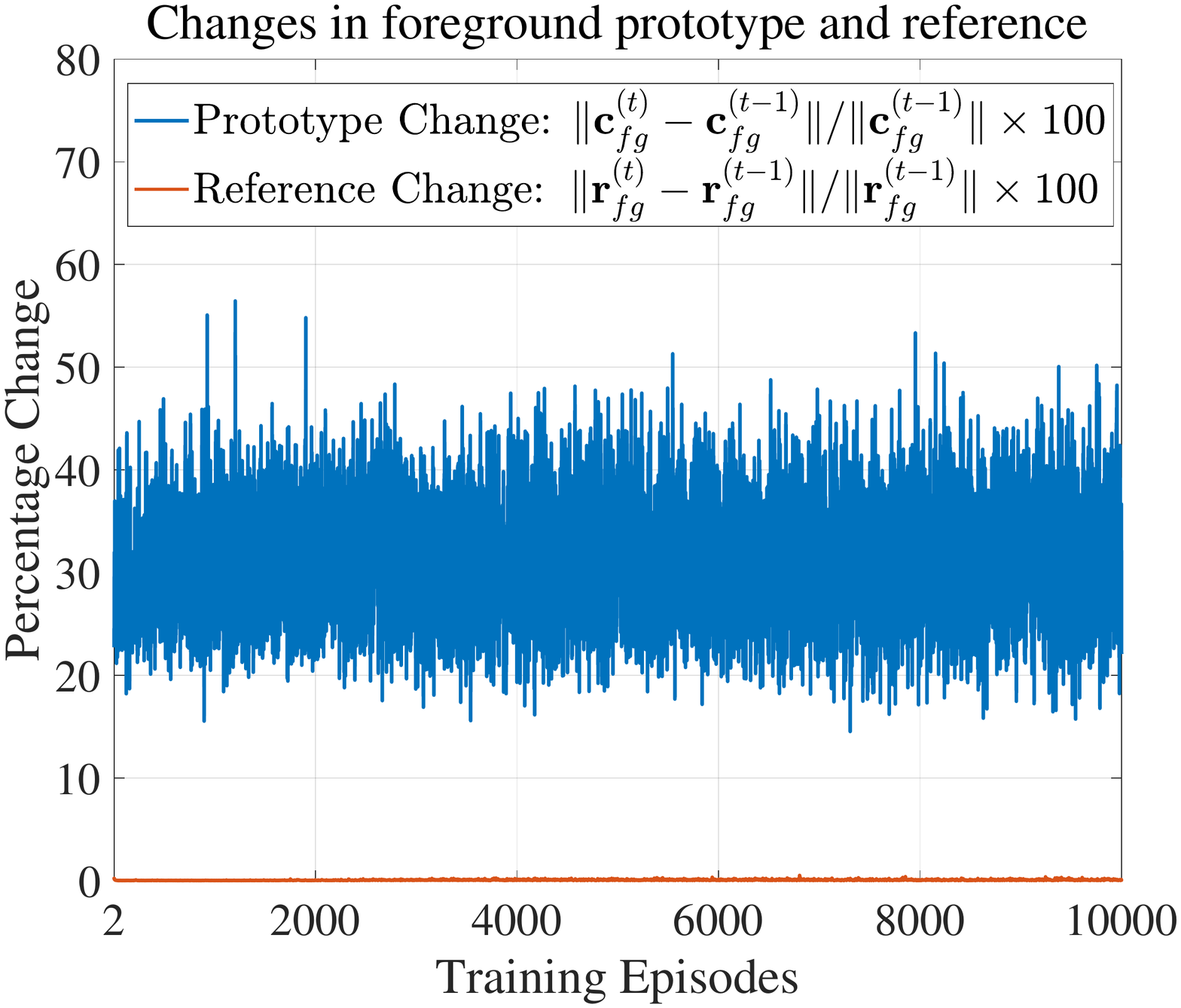}
			\subcaption{Foreground class}
		\end{subfigure}	
		\begin{subfigure}[b]{0.48\textwidth}
			\includegraphics[width=\textwidth]{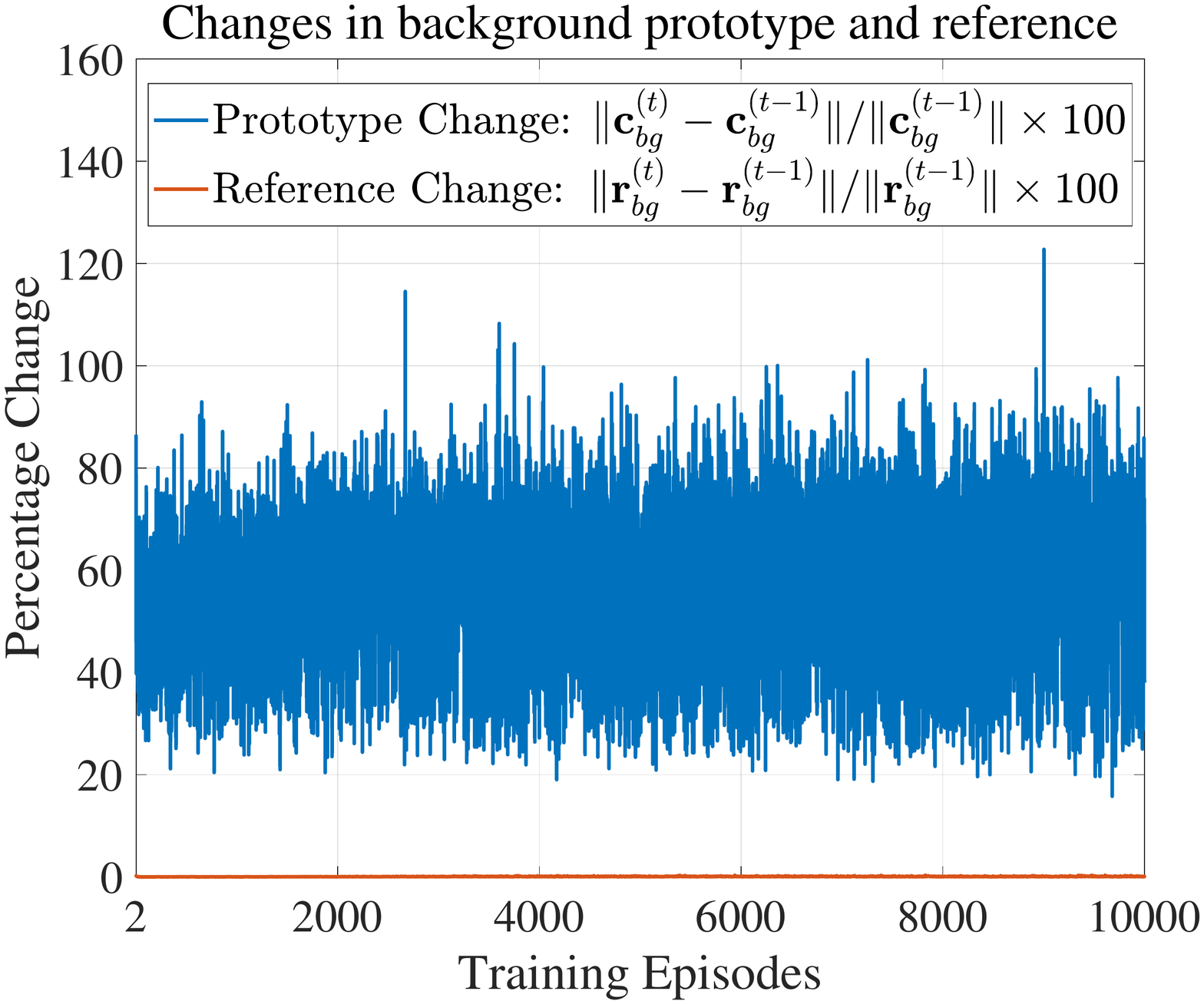}
			\subcaption{Background class}
		\end{subfigure}	
		\caption{Comparison of changes in prototypes and references during meta-training}
		\label{figure_app_2}
	\end{figure*}

	In Section \ref{section2_1}, we stated that the prototypes $\mathbf{c}_k$ change significantly while the reference vectors $\mathbf{r}_k$ changes only by little. To confirm this, we collect the prototypes and the reference vectors of foreground and background classes from the first episode to the 10,000th episode during meta-training. Then we compute the Euclidean distances $\norm{\mathbf{c}^{(t)}_{k}-\mathbf{c}^{(t-1)}_{k}}$ and  $\norm{\mathbf{r}^{(t)}_{k}-\mathbf{r}^{(t-1)}_{k}}$ for $k\in\{fg, bg\}$ and $t \in [2, 10000]$ to measure the changes. In Figure \ref{figure_app_2}, we visualize the percentage changes of the prototypes and reference vectors during meta-training. To focus on the reference vector in TAFT module, we utilize TAFT on Deeplab V3+ without SE module for this experiment. We adopt the ResNet-50 network as the backbone encoder. We can see that the prototypes undergo significant changes while the relative changes in the reference vectors are nearly invisible. Although the class in each episode is randomly sampled and the class may or may not change from one episode to the next, the changes in prototype are always much larger than the changes in reference vectors, with or without class changes. 
	
	\subsection{Ablation Study on Transforming Low Level Feature}

	In the experiments on TAFT-SE on Deeplab V3+ in the main paper, we do not apply the feature transformation via TAFT to the low-level feature, since it contains shape information which can be considered general. In Tables \ref{table8} and \ref{table9}, we explore the effect of applying feature transformation to the low-level feature of Deeplab V3+. We apply the low-level feature transformation by adding another reference vector set for low-level feature, and training it using additional regression loss. The transformation matrix is constructed using the additional reference vectors and the prototypes computed from low-level support features. In order to more clearly understand the effect of low-level feature transformation, we do not utilize the SE module in this experiment again; we use the TAFT on Deeplab V3+ model. The results in Tables \ref{table8} and \ref{table9} show that applying feature transformation at low-level feature does not bring any gain in performance; it actually degrades 1-shot and 5-shot mIoU by 1.39 and 0.74, and 1-shot and 5-shot FBIoU by 0.70 and 0.61.
	\begin{table*}[h!]
		\centering
		\scriptsize
		\begin{tabular}{c|c|c|c|c|c}
			\toprule  
			\textbf{Models}    & split-0& split-1	&split-2 &split-3 &mean \\
			\midrule
			\textbf{TAFT on Deeplab V3+} \small{(1-shot, w/ low-level TAFT)}   &45.59	&56.98&50.01 &47.48&50.02 \\
			\textbf{TAFT on Deeplab V3+} \small{(1-shot)}   &48.00	&57.64&52.75 &47.26&51.41  \\
			\textbf{TAFT on Deeplab V3+} \small{(5-shot, w/ low-level TAFT)}   &55.91	&67.88&62.93 &58.86&61.40 \\
			\textbf{TAFT on Deeplab V3+} \small{(5-shot)}   &57.36	&68.03&63.71 &59.45&62.14  \\
			\bottomrule
		\end{tabular}
		\caption{PASCAL-$5^i$ mIoU of TAFT on Deeplab V3+ with and without low-level feature transformation}
		\label{table8}
	\end{table*}
	
	\begin{table*}[h!]
		\centering
		\scriptsize
		\begin{tabular}{c|c|c|c|c|c}
			\toprule  
			\textbf{Models}    & split-0& split-1	&split-2 &split-3 &mean \\
			\midrule
			\textbf{TAFT on Deeplab V3+} \small{(1-shot, w/ low-level TAFT)}   &66.82	&74.27&67.83 &66.75&68.91 \\
			\textbf{TAFT on Deeplab V3+} \small{(1-shot)}  &67.00	&75.12&69.44 &66.87&69.61  \\
			\textbf{TAFT on Deeplab V3+} \small{(5-shot, w/ low-level TAFT)}   &72.07	&79.81&74.90 &74.70&75.37\\
			\textbf{TAFT on Deeplab V3+} \small{(5-shot)}   &72.94	&80.15&75.49 &75.35&75.98  \\
			\bottomrule
		\end{tabular}
		\caption{PASCAL-$5^i$ FBIoU of TAFT on Deeplab V3+ with and without low-level feature transformation}
		\label{table9}
	\end{table*}
	
	\section{Additional Qualitative Results}
\begin{figure*}[h]
		\centering
		\includegraphics[width=0.9\textwidth]{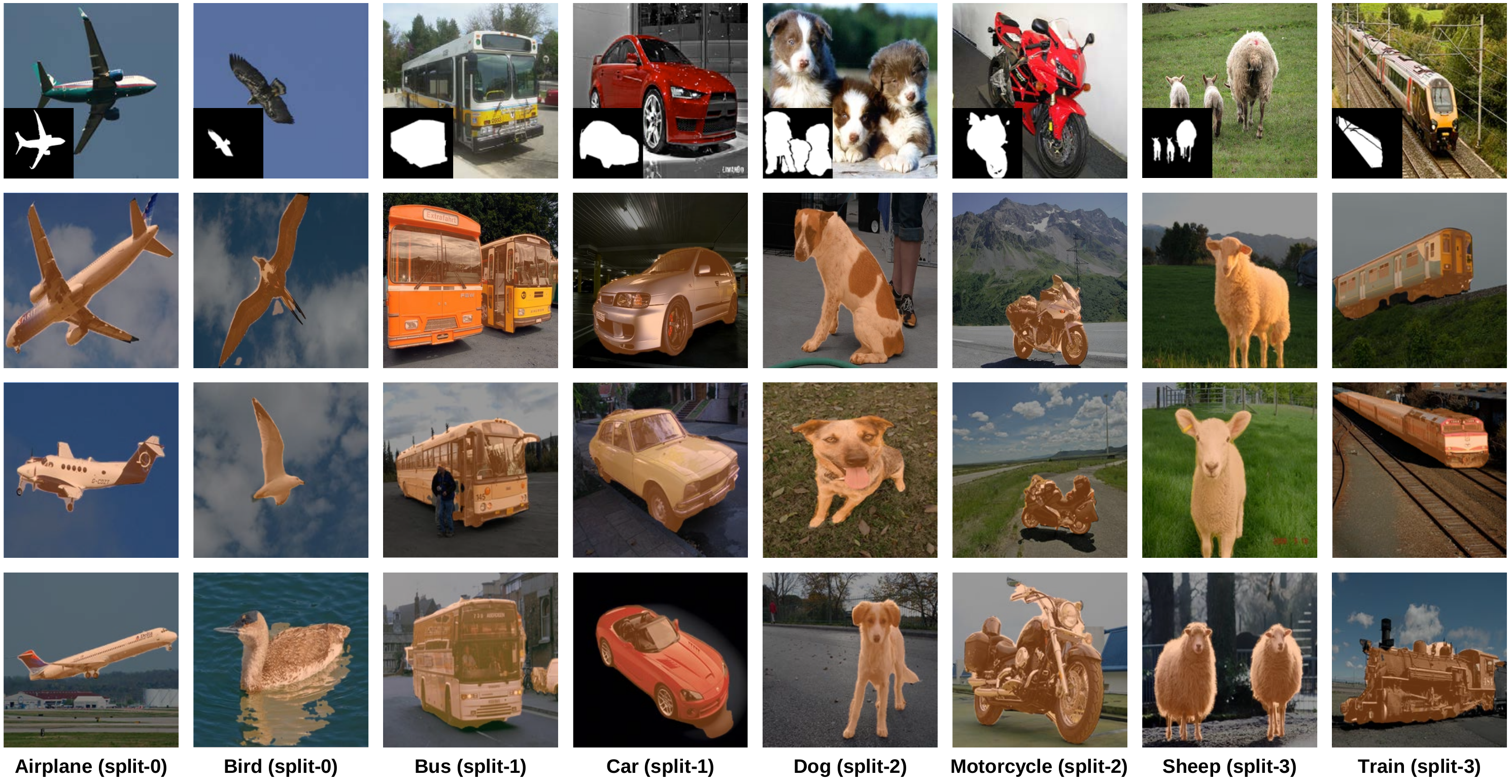}
		\caption{Qualitative results of TAFT-SE on Deeplab V3+ for 1-shot segmentation on PASCAL-$5^i$ dataset}
		\vspace{-5mm}
		\label{figure_app_3}
	\end{figure*}

\begin{figure*}[h]
		\centering
		\includegraphics[width=0.9\textwidth]{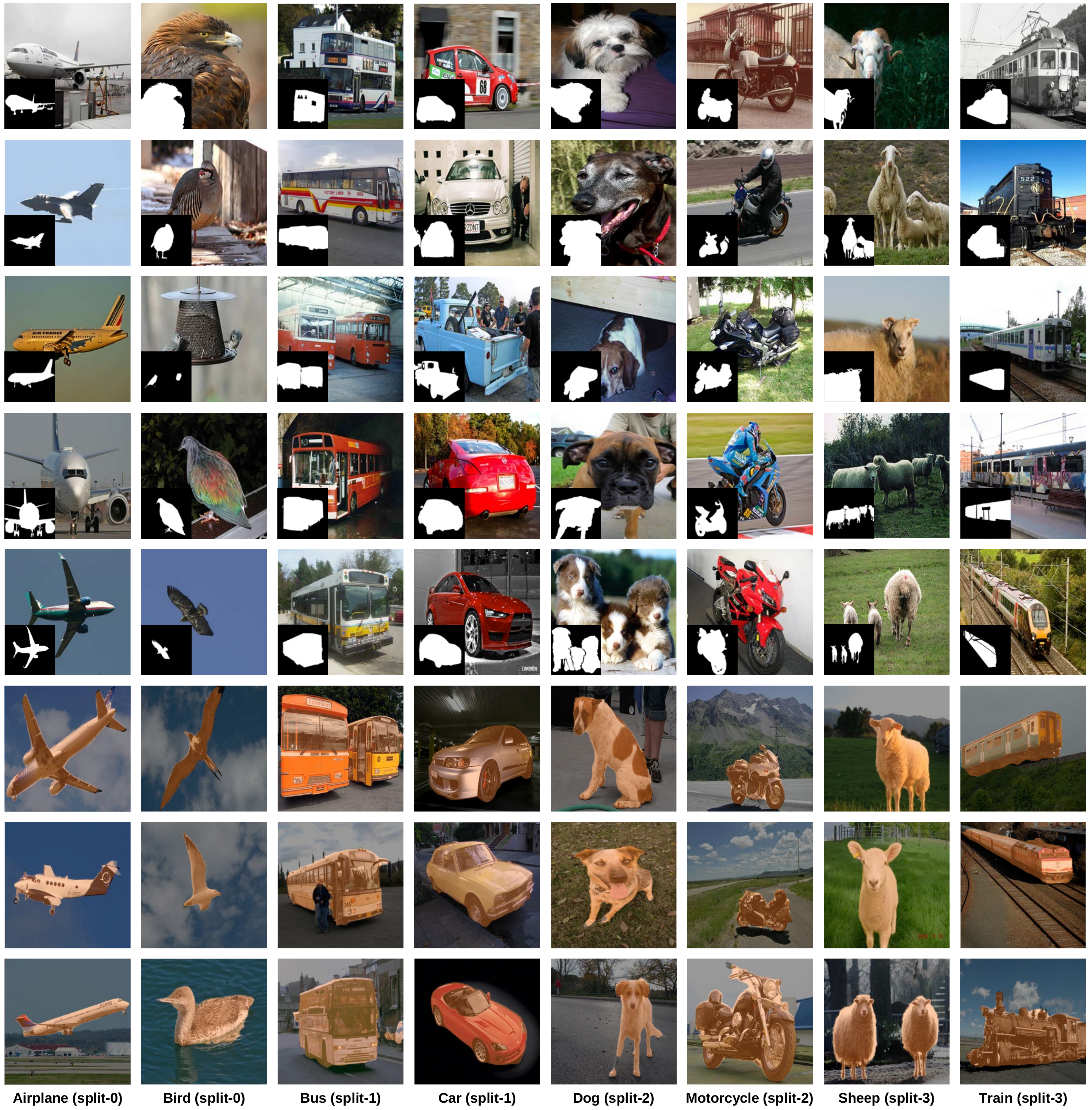}
		\caption{Qualitative results of TAFT-SE on Deeplab V3+ for 5-shot segmentation on PASCAL-$5^i$ dataset}
		\vspace{-5mm}
		\label{figure_app_4}
	\end{figure*}

	In Figures \ref{figure_app_3} and \ref{figure_app_4}, we display additional qualitative results of TAFT-SE on Deeplab V3+. The 1-shot and 5-shot segmentation results of airplane, bird, bus, car, dog, motorcycle, sheep and train classes are visualized. As done in the main paper, we show the support set images with the support labels together at the bottom left, and the query images with the prediction results. From these results, we can observe that the quality of segmentation is improved from 1-shot to 5-shot.    
	
\end{document}